%% file: sample.tex
\documentclass[twoside,11pt]{article}

\usepackage{amsmath}
\usepackage{enumitem}
\usepackage{amssymb}
\usepackage{mathtools}
\usepackage{amsthm}
\usepackage{bm}
\usepackage{color}
\usepackage{multirow}
\usepackage{subcaption}
\usepackage{booktabs}
\usepackage{url}
\usepackage{graphicx}
\usepackage{algorithm}
\usepackage{algorithmic}
\usepackage{amsfonts}
\usepackage{blindtext}

\usepackage{jmlr2e}

\usepackage{lastpage}

\ShortHeadings{NDCG-Consistent Softmax Approximation with Accelerated Convergence}{Pu, Lian, Chen, Huang, Chen and Chen}
\firstpageno{1}

\begin{document}

\title{NDCG-Consistent Softmax Approximation with Accelerated Convergence}

\author{\name Yuanhao Pu \email puyuanhao@mail.ustc.edu.cn \\
       \addr School of Artificial Intelligence \& Data Science\\
       University of Science and Technology of China
       \AND
       \name Defu Lian\thanks{Corresponding author.} \email liandefu@ustc.edu.cn \\
       \addr School of Computer Science \& Technology\\
       University of Science and Technology of China
       \AND
       \name Xiaolong Chen \email chenxiaolong@mail.ustc.edu.cn \\
       \addr School of Artificial Intelligence \& Data Science\\
       University of Science and Technology of China
       \AND
       \name Xu Huang \email xuhuangcs@mail.ustc.edu.cn \\
       \addr School of Computer Science \& Technology\\
       University of Science and Technology of China
       \AND
       \name Jin Chen \email jinchen@ust.hk \\
       \addr School of Business \& Management\\
       Hong Kong University of Science and Technology
       \AND
       \name Enhong Chen \email cheneh@ustc.edu.cn\\
       \addr School of Computer Science \& Technology\\
       University of Science and Technology of China}

\editor{My editor}

\maketitle

\begin{abstract}
Ranking tasks constitute fundamental components of extreme similarity learning frameworks, where extremely large corpora of objects are modeled through relative similarity relationships adhering to predefined ordinal structures. Among various ranking surrogates, Softmax (\textbf{SM}) Loss has been widely adopted due to its natural capability to handle listwise ranking via global negative comparisons, along with its flexibility across diverse application scenarios. However, despite its effectiveness, \textbf{SM} Loss often suffers from significant computational overhead and scalability limitations when applied to large-scale object spaces. To address this challenge, we propose novel loss formulations that align directly with ranking metrics: the \textbf{R}anking-\textbf{G}eneralizable \textbf{squared} (\textbf{RG$^2$}) Loss and the \textbf{R}anking-\textbf{G}eneralizable \textbf{interactive} (\textbf{RG$^\times$}) Loss, both derived through Taylor expansions of the \textbf{SM} Loss. Notably, \textbf{RG$^2$} reveals the intrinsic mechanisms underlying weighted squared losses (\textbf{WSL}) in ranking methods and uncovers fundamental connections between sampling-based and non-sampling-based loss paradigms. Furthermore, we integrate the proposed \textbf{RG} losses with the highly efficient Alternating Least Squares (\textbf{ALS}) optimization method, providing both generalization guarantees and convergence rate analyses. Empirical evaluations on real-world datasets demonstrate that our approach achieves comparable or superior ranking performance relative to \textbf{SM} Loss, while significantly accelerating convergence. This framework offers the similarity learning community both theoretical insights and practically efficient tools, with methodologies applicable to a broad range of tasks where balancing ranking quality and computational efficiency is essential.

\end{abstract}


\input{section/intro}

\input{section/related.tex}

\input{section/preliminary}

\input{section/method}

\input{section/exp}

\input{section/conclusion}




\newpage

\appendix

\section{Theoretical Results}

\subsection{Calibration and Consistency of Softmax Loss via Bregman Divergence}
\label{Appdix.SMConsistent}
To clarify the theoretical guarantees of \textbf{SM} under both classification and ranking tasks, we separately discuss its desirable properties in \textbf{Top-$k$ classification} and \textbf{ranking with DCG-based metrics}, highlighting that both analyses share a unified foundation through \textbf{Bregman Divergence}. In line with the discussions in \cite{yang2020consistency} and \cite{ravikumar2011ndcg}, we adopt the term \emph{calibration} for classification and \emph{consistency} for ranking, acknowledging that these notions are deeply connected via conditional optimality in the infinite sample regime.

\subsubsection{Top-$k$ Calibration of Softmax Loss}

The concept of \textbf{Top-$k$ calibration} characterizes the ability of a surrogate loss to recover the Bayes-optimal decision under the Top-$k$ selection rule. We first introduce the notions of \textbf{Top-$k$ preserving} and \textbf{inverse Top-$k$ preserving functions}.

\begin{definition}[Top-$k$ preserving]
Given vectors $\bm{s}, \bm{\eta} \in \mathbb{R}^N$, we say that $\bm{s}$ is top-$k$ preserving with respect to $\bm{\eta}$, denoted as $P_k(\bm{s}, \bm{\eta})$, if for all $n \in [N]$,
\[
\eta_n > \eta_{[k+1]} \implies s_n > s_{[k+1]},
\quad
\eta_n < \eta_{[k]} \implies s_n < s_{[k]},
\]
where $s_{[j]}$ denotes the $j$-th largest element in $\bm{s}$.
\end{definition}

\begin{definition}[Inverse Top-$k$ preserving function]
A function $f: A \to B$ is said to be inverse top-$k$ preserving if for all $\bm{s} \in A$, $P_k(\bm{s}, f(\bm{s}))$ holds.
\end{definition}

The formal definition of \textbf{Top-$k$ calibration} for surrogate losses is as follows:

\begin{definition}[Top-$k$ calibration]
Let $\Delta_N = \{ \bm{\eta} \in \mathbb{R}^N \mid \sum_{i=1}^N \eta_i = 1 \}$ be the probability simplex. A loss function $\psi: \mathbb{R}^N \times \mathcal{Y} \to \mathbb{R}$ is top-$k$ calibrated if for all $\bm{\eta} \in \Delta_N$,
\[
\inf_{\bm{s} \in \mathbb{R}^N \cap \lnot P_k(\bm{s}, \bm{\eta})} L_\psi(\bm{s}, \bm{\eta}) > \inf_{\bm{s} \in \mathbb{R}^N} L_\psi(\bm{s}, \bm{\eta}) = L_\psi(\bm{\eta}),
\]
where $L_\psi(\bm{s}, \bm{\eta})$ denotes the conditional risk under distribution $\bm{\eta}$.
\end{definition}.

\subsubsection{DCG-Consistency of Softmax Loss}

Beyond classification, \textbf{SM} also demonstrates desirable properties in \textbf{ranking tasks}. Following \cite{ravikumar2011ndcg}, we adopt the notion of \textbf{DCG-consistency}, which ensures that the surrogate loss, when minimized, recovers the Bayes-optimal ranking under \textbf{DCG}.

To support this, we introduce the concept of \textbf{order preserving}:

\begin{definition}[Order preserving]
Given vectors $\bm{s}, \bm{\eta} \in \mathbb{R}^N$, we say that $\bm{s}$ is order preserving with respect to $\bm{\eta}$, denoted as $\bm{s} \hookrightarrow \bm{\eta}$, if
\[
\forall i, j, \ \eta_i > \eta_j \implies s_i > s_j.
\]
A function $g(\cdot)$ is order preserving if $g(\bm{s}) \hookrightarrow \bm{s}$ holds for all $\bm{s}$.
\end{definition}

\cite{ravikumar2011ndcg} demonstrated that the \emph{inverse order preserving} property of the softmax map, together with the strict convexity of its generating function $\phi$, guarantees the \textbf{DCG-consistency} of \textbf{SM} (as shown in Prop.\ref{prop.SMConsist}). This result also relies on the Bregman divergence framework.

\begin{remark}[Normalization in surrogate losses] \cite{ravikumar2011ndcg} emphasized that normalization of relevance scores is required for $\operatorname{NDCG}$-consistency. However, in large-scale implicit feedback settings where relevance is typically binary (single-click behavior), normalization becomes trivial as all weights reduce to one, consistent with the argument in \cite{bruch2021alternative}. Throughout this paper, we focus on $\operatorname{DCG}$-consistency without explicit normalization to avoid unnecessary complexity.
\end{remark}

\subsection{Transformation of RG$^2$ Realization}\label{Appdix.transformation}

\begin{align*}
    \mathcal{L}_{\operatorname{RG^2}} &= \mathbb{E}_{\mathcal{D}}\left[-o_y^{(x)}+\frac{1}{2N}\|\bm{o}^{(x)}+\bm{1}_N\|^2\right]+\lambda\psi(\theta)\\
    &=\frac{1}{|D|}\sum_{(x,y)\in D}\left(-o^{(x)}_{y}+\frac{1}{2N}\|\bm{o}^{(x)}+\bm{1}_N\|^2\right)+\lambda\psi(\theta)\quad\text{(Realization)}\\
    &= \frac{1}{|D|}\sum_{(x,y)\in D}\left(-o_y^{(x)}+\sum_{y'=1}^N\frac{1}{2N}(o_{y'}^{(x)}+1)^2\right)+\lambda\psi(\theta)\\
    &= \frac{1}{|D|}\left(-\sum_{(x,y)\in D}o_y^{(x)}+\sum_{x\in X}|\mathcal{I}_x|\sum_{y'=1}^N\frac{1}{2N}(o_{y'}^{(x)}+1)^2\right)+\lambda\psi(\theta) \\
    &= \frac{1}{|D|}\left(\sum_{(x,y)\in D}\left(\frac{|\mathcal{I}_x|}{2N}(o_y^{(x)}+1)^2-o_y^{(x)}\right)+\sum_{(x,y)\notin D}\frac{|\mathcal{I}_x|}{2N}(o_y^{(x)}+1)^2\right)+\lambda\psi(\theta) \\
    &= \frac{1}{|D|}\left(\sum_{(x,y)\in D}\frac{|\mathcal{I}_x|}{2N}\left({o_y^{(x)}}^2+2o_y^{(x)}-\frac{2N}{|\mathcal{I}_x|}o_y^{(x)}+1\right)+\sum_{(x,y)\notin D}\frac{|\mathcal{I}_x|}{2N}(o_y^{(x)}+1)^2\right)+\lambda\psi(\theta) \\
    &\propto \frac{1}{|D|}\left(\sum_{(x,y)\in D}\frac{|\mathcal{I}_x|}{2N}\left(\left(o_y^{(x)}+1-\frac{N}{|\mathcal{I}_x|}\right)^2\right)+\sum_{(x,y)\notin D}\frac{|\mathcal{I}_x|}{2N}(o_y^{(x)}+1)^2\right)+\lambda\psi(\theta) \\
    &= \frac{1}{|D|} \sum_{x\in X,y\in Y}  \frac{|\mathcal{I}_x|}{2N} \left(o_y^{(x)}+1 - r_{x,y}\frac{N} {|\mathcal{I}_x|} \right)^2 + \lambda\psi(\theta)\\
    &\propto \frac{1}{|D|} \sum_{x\in X,y\in Y}  |\mathcal{I}_x| \left(o_y^{(x)}+1 - r_{x,y}\frac{N} {|\mathcal{I}_x|} \right)^2 + \lambda\psi(\theta) \quad \text{(Absorb Coefficient)}
\end{align*}

\subsection{Proof of SM Upper Bound}\label{Appdix.UpperBound}
To show that the approximation in Eq.(\ref{eq.RGsquaredform}) is an upper bound of SM Loss, consider the following lemma:
\begin{lemma}
    For any multi-variable function $f:\mathbb{R}^N\to\mathbb{R}$ with Jacobian vector $\nabla f$ and Hessian matrix $\nabla^2 f$, let $M$ be a symmetric matrix satisfying $A\succeq \nabla^2 f$, then for $\forall x,y\in\mathbb{R}^N$,
    \begin{equation*}
        f(y)\leq f(x) + \nabla f(x)(y-x) + \frac{1}{2}(y-x)^\top A (y-x)
    \end{equation*}
    \begin{proof}
        With the mean value theorem, one can simply expand $f(y)$ with
        \begin{equation*}
            f(y) = f(x) + \nabla f(x)(y-x) + \frac{1}{2}(y-x)^\top\nabla^2 f(\xi)(y-x)
        \end{equation*}
        Now that $A\succeq \nabla^2 f$, thus for $\forall x\in\mathbb{R}^N, x^\top A x\geq x^\top\nabla^2 f(\xi) x$, hence finishing the proof.
    \end{proof}
\end{lemma}

To give out an upper bound of \textbf{SM}, we claim that symmetric matrix $\frac{1}{N}I_N-\nabla^2f$ is PSD under single-click assumption.
\begin{proof}[Proposition \ref{prop.upperbound}]
    The proof reduces to verifying whether the symmetric matrix
    \[
    A = \frac{1}{N} I_N - \operatorname{diag}(\bm{p}) + \bm{p} \bm{p}^\top
    \]
    is positive semi-definite, where $\bm{p} = [p(\bm{o}^{(x)}_1),\cdots,p(\bm{o}^{(x)}_N)]$.

    To guarantee $A \succeq 0$, we analyze its behavior along the critical direction $\bm{p}$, which corresponds to the potential minimum eigenvector of the rank-one update $\bm{p} \bm{p}^\top$. The Rayleigh quotient of $A$ along $\bm{p}$ is given by
    \[
    R(\bm{p}) 
    = \frac{\bm{p}^\top A \bm{p}}{\|\bm{p}\|^2} 
    = \frac{1}{N} - \frac{\sum_{j=1}^N p_j^3}{\sum_{j=1}^N p_j^2} + \sum_{j=1}^N p_j^2.
    \]
    Hence, a sufficient condition for $A \succeq 0$ is
    \begin{equation}\label{eq:concentration_condition}
        \frac{1}{N} - \frac{\sum_{j=1}^N p_j^3}{\sum_{j=1}^N p_j^2} + \sum_{j=1}^N p_j^2 \geq 0.
    \end{equation}

    Noting that $\sum_{j=1}^N p_j^3 \leq p_{\max} \sum_{j=1}^N p_j^2$, where $p_{\max} = \max_j p_j$, we obtain a relaxed sufficient condition:
    \[
    p_{\max} \leq \frac{1}{N} + \sum_{j=1}^N p_j^2.
    \]
    This inequality is satisfied when $p$ is either uniformly distributed (our initial setting $\bm{o}_0=\bm{0}$, where $p_j = 1/N$ for all $j$) or highly concentrated on a single entry ($p_{\max}=\frac{\exp(N-1)}{\exp(N-1)+(N-1)\exp(-1)}$ $\to 1$ and $p_j=\frac{\exp(-1)}{\exp(N-1)+(N-1)\exp(-1)} \to 0$ for $j \neq \arg\max p_j$), corresponding to the single-click behavior assumption. This completes the proof.
\end{proof}

\subsection{Proof of Lemma \ref{lem.dcgconsist}}\label{pf.lem6}

\begin{proof}[Lemma \ref{lem.dcgconsist}]
    \begin{align*}
        \mathcal{L}_{\text{DCG}}(\bm{\pi}) &= \mathbb{E}_{\mathcal{D}}\left[- \sum_{y=1}^N c_{\pi_y}r_{x,y}\right] = -\sum_{y=1}^N c_{\pi_y} f_B(x,y)\\
        &=-\sum_{y=1}^N c_{\pi_y} f(x,y)-\sum_{y=1}^N c_{\pi_y}\left(f_B(x,y)-f(x,y)\right)\\
        &\leq -\sum_{y=1}^N c_{\pi_y^*} f(x,y)-\sum_{y=1}^N c_{\pi_y}\left(f_B(x,y)-f(x,y)\right)\\
        &= -\sum_{y=1}^N c_{\pi_y^*} f_B(x,y)-\sum_{y=1}^N c_{\pi_y^*}\left(f(x,y)-f_B(x,y)\right)-\sum_{y=1}^N c_{\pi_y}\left(f_B(x,y)-f(x,y)\right)\\
        &= \mathcal{L}_{\text{DCG}}^*-\sum_{y=1}^N c_{\pi_y^*}\left(f(x,y)-f_B(x,y)\right)-\sum_{y=1}^N c_{\pi_y}\left(f_B(x,y)-f(x,y)\right)\\
        &\leq \mathcal{L}_{\text{DCG}}^* + \left(2\sum_{y=1}^N c_y^2\right)^{\frac{1}{2}}\left(\sum_{y=1}^N\left(f(x,y)-f_B(x,y)\right)^2\right)^{\frac{1}{2}}
    \end{align*}
\end{proof}

\subsection{Proof of Generalization Upper Bound}
\label{Appdix.GUpperBound}

\paragraph{Pseudo-Dimension of the Hypothesis Space.}
The following result provides an upper bound on the pseudo-dimension of low-rank matrix classes:
\begin{proposition}[\cite{srebro2004generalization}]
The pseudo-dimension of the low-rank matrix hypothesis space $\mathcal{H}$ is at most:
\[
d = K(M + N) \log \frac{16eM}{K},
\]
where $M, N, K$ represent the number of contexts, objects, and embedded dimensions.
\end{proposition}

\paragraph{Covering Number and Generalization Bound.}
According to ~\cite{anthony1999neural}, the $L_1$-covering number of any hypothesis class $\mathcal{H}$ with finite pseudo-dimension $d$ satisfies:
\[
\mathcal{N}_1(\varepsilon, \mathcal{H}, m) < e(d + 1) \left( \frac{2e}{\varepsilon} \right)^d.
\]

With the Lipschitz condition of the loss function, $L_1$-covering number of hypothesis space $\mathcal{H}$ establishes a direct relation to the generalized error bound, from which obtaining the following theorem.

\begin{lemma}~\citep{anthony1999neural}\label{lem:general upper bound}
     Suppose $\mathcal{H}$ be a nonempty set of real functions mapping from a domain $\mathcal{X}$ into $[0,1]$. Let $P$ be any probability distribution on $\mathcal{X}\times\mathbb{R}$, $\forall\varepsilon > 0$ with any positive integer $m$. Then for any loss function $\ell(h(x),y)$ with $|\ell(h(x),y)|\leq B$ with $L$-Lipschitz,
     \begin{equation}
     \begin{aligned}
         P^m\left(\exists h\in\mathcal{H}\quad s.t.\quad |\mathcal{R}(h)-\hat{\mathcal{R}}_{\mathcal{D}}(h)|\geq\varepsilon\right)\leq 4\mathcal{N}_1\left(\frac{\varepsilon}{8L},\mathcal{H}, 2m\right)\exp\left(\frac{-m\varepsilon^2}{32B^4}\right)
     \end{aligned}\end{equation}
\end{lemma}

We have listed all the conditions required to prove an upper bound on generalization error. By combining the above conclusions, the proof of Theorem.\ref{Thm:GUpperBound} is given as follows:

\begin{proof}[Theorem.\ref{Thm:GUpperBound}]
    Starting from Lemma~\ref{lem:general upper bound}, we have
    \begin{equation}\begin{aligned}
    &P^m\left(|\mathcal{R}(h)-\hat{\mathcal{R}}_{\mathcal{D}}(h)|\geq\varepsilon\right)\leq 4\mathcal{N}_1\left(\frac{\varepsilon}{8L},\mathcal{H}, 2m\right)\exp\left(-\frac{|\mathcal{D}|\varepsilon^2}{32B^4}\right)\\
    &\leq 4e(d+1)(\frac{32eB}{\varepsilon})^d\exp\left(-\frac{|\mathcal{D}|\varepsilon^2}{32B^4}\right)\leq 4e(d+1)(\frac{16eL}{\varepsilon})^d\frac{32B^4}{|\mathcal{D}|\varepsilon^2}
    \end{aligned}\end{equation}
    The last inequality comes from $\exp(-x)\leq\frac{1}{x}$. By setting
    \begin{equation}
        \delta = 4e(d+1)(\frac{16eL}{\varepsilon})^d\frac{32B^4}{|\mathcal{D}|\varepsilon^2},
    \end{equation}
    we obtain the final form of $\varepsilon$ in Eq.(\ref{eq.generalization bound}).
\end{proof}

\subsection{Advanced ALS Optimization through matrix operation}\label{appdix.A-ALS}
Consider the \textbf{RG$^2$} formulation in Eq.(\ref{eq:RG2 MF form}), we have
\begin{equation}
    \begin{aligned}
        \mathcal{L}_{\text{RG$^2$}} &=\sum_{x,y=1}^{M,N}W_{x}(S_{x,y}-P_{x\cdot}\cdot Q_{\cdot y}^\top)^2+\lambda(\|P\|_F^2+\|Q\|_F^2)\\
        &=\operatorname{tr}\left((S-PQ^\top)^\top W(S-PQ^\top)\right)+\lambda\cdot\operatorname{tr}(P^\top P+Q^\top Q)
    \end{aligned}
\end{equation}
where $W^{m\times m}=\operatorname{diag}(W_x)$. It is intuitive to find the gradient of the obtained matrix equation with respect to $P$ and $Q$. Unfortunately, directly solving for the gradient of this form produces a \textit{sylvester equation} with respect to $P$, which is not a desired result.
\begin{equation}
\begin{aligned}
    \nabla_P\mathcal{L}&=-2WSQ+2WPQ^\top Q+2\lambda P = 0\\
    &\implies WPQ^\top Q+\lambda P=WRQ
\end{aligned}
\end{equation}
To avoid this problem and obtain closed-form solutions for $P$ and $Q$ directly, an acceptable modification would be to change the regularization term from $\lambda(\|P\|_F^2+\|Q\|_F^2)$ into $\lambda(\|PQ^\top\|_F^2)$, which is also widely chosen in many practices. Then
\begin{equation}
    \begin{aligned}
        \nabla_P\mathcal{L}&=-2WSQ+2WPQ^\top Q+2\lambda PQ^\top Q = 0\\
        & \implies P = (W+\lambda I)^{-1}WSQ(Q^\top Q)^{-1}\\
        \nabla_Q\mathcal{L}&=-2S^\top WP+2QP^\top WP+2\lambda QP^\top P = 0\\
        & \implies Q = S^\top WP(\lambda P^\top P +P^\top W P)^{-1}\\
    \end{aligned}
\end{equation}
From this, we can modify the optimization method in Algorithm \ref{alg:wALS} with the update of full matrix rather than row-by-row.

\subsection{Analysis of Convergence Rate}\label{Appdix.ConvergRate}

\textbf{SGD} remains a fundamental optimization algorithm for nonconvex problems. Under standard assumptions of smoothness and bounded variance, \textbf{SGD} achieves a convergence rate towards stationary points characterized as follows:
\begin{equation}
\mathbb{E} \left[ \|\nabla \mathcal{L}(P, Q)\|^2 \right] \leq \mathcal{O}\left( \frac{1}{\sqrt{T}} \right),
\end{equation}
where $T$ denotes the total number of stochastic updates. Such behavior has been formally established in the literature on nonconvex optimization \citep{ghadimi2013stochastic, bottou2018optimization}. Without specialized variance reduction techniques\citep{johnson2013accelerating, defazio2014saga}, the global convergence rate of standard \textbf{SGD} on the joint variable space cannot be improved.

\textbf{Newton-CG} is a classical second-order optimization method designed to solve large-scale nonconvex problems by approximating Newton steps via conjugate gradient iterations. Under the second-order sufficient conditions and sufficiently accurate solutions of the Newton system, \textbf{Newton-CG} enjoys a local superlinear convergence rate towards a strict local minimum \citep{dembo1982inexact}. These properties make \textbf{Newton-CG} particularly appealing when high-precision solutions are required and saddle points can be effectively avoided.




\textbf{ALS} is a deterministic optimization method commonly used for non-convex problems, particularly those exhibiting marginal convexity structures, such as matrix factorization tasks arising in recommendation systems. It alternates between updating left and right object embeddings while holding the other fixed, exploiting the block-wise convexity to achieve effective progress. Under standard assumptions, \textbf{ALS} exhibits a global linear convergence rate towards a stationary point, formalized as follows:

\begin{theorem}[\cite{jain2017non}]
Suppose the loss function $\mathcal{L}(P, Q)$ is continuously differentiable and satisfies the properties of $\mu$-marginal strong convexity (MSC) and $L$-marginal strong smoothness (MSS) in both $P$ and $Q$ individually, and further possesses $M$-robust bistability. Then, the ALS algorithm satisfies:
\begin{equation}
|\mathcal{L}(P_T, Q_T) - \mathcal{L}^*| \leq \left(\frac{2L(M+\mu)}{\mu^2+2L(M+\mu)}\right)^T |\mathcal{L}(P_0, Q_0) - \mathcal{L}^*| \sim \mathcal{O}(\rho^T),
\end{equation}
where $\rho = \frac{2L(M+\mu)}{\mu^2+2L(M+\mu)} < 1$ denotes the contraction factor per iteration.
\end{theorem}

It is important to note that in our setting, the \textbf{RG} losses, although globally nonconvex, satisfy marginal strong convexity and smoothness properties due to their quadratic structures, ensuring the applicability of the above convergence theorem.

\section{Experimental Details}\label{appdix.exp}

\subsection{Description of Datasets}\label{appdix.data}

\textbf{MovieLens} comprises approximately 10 million movie ratings ranging from 0.5 to 5, in increments of 0.5. \textbf{Electronics} collects the customer's reviews on electronics products on the Amazon platform, where each review consists of a rating ranging from 0 to 5 and the reviews about the product. \textbf{Steam} is a dataset crawled from the large online video game distribution platform \textit{Steam}(\cite{kang2018self}), comprising the player's reviews plus rich information such as playing hours. \textbf{Wiki} is a hyperlink network dataset derived from the full revision history of Simple English Wikipedia articles\footnote{https://dumps.wikimedia.org/simplewiki/20250101/}. It contains approximately 20 million directed links between articles, where each entry encodes a source-to-target page connections. 
As for \textbf{MovieLens} and \textbf{Electronics}, we treat items rated below 3 as negatives and the remains as positives. For \textbf{Steam}, since there is no explicit rating, we treat all samples as positives. For \textbf{Wiki}, we treat all page$_{\text{from}}$ and page$_{\text{to}}$ id pairs as positive left and right components. We employ the widely used k-core filtering strategy to filter out the users and items with interactions less than 5 for all recommendation datasets and 20 for \textbf{Wiki}. The detailed statistics of those datasets after filtering are illustrated in Table~\ref{tab:dataset_stas}.

\textbf{Data Split.} We partition all datasets into training, validation, and test sets with a split ratio of $\{0.8, 0.1, 0.1\}$ for each user, respectively. The validation set is utilized to assess the model's performance, while the metrics derived from the test set serve as the foundation for comparative analysis.

\begin{table}[thb]
    \centering
    \caption{Statistics of datasets.}
    \begin{tabular}{l|r|r|r|r}
        \toprule
         Dataset & \#Left & \#Right & \#Interact & Sparsity  \\ \midrule
         MovieLens & 69,815 & 9,888 & 8,240,192 & 98.81\% \\ \midrule
         Electronics & 192,403 & 63,001 & 1,689,188 & 99.99\% \\ \midrule
         Steam & 281,204 & 11,961 & 3,484,497 & 99.90\% \\ \midrule     
         Wiki & 129,404 & 129,432 & 13,715,724 & 99.92\%\\
         \bottomrule
    \end{tabular}
    \label{tab:dataset_stas}
\end{table}

\subsection{Description of Baselines}
\begin{itemize}[leftmargin=*]
    \item \textbf{BPR}~\citep{rendle2012bpr}: Bayesian Personalized Ranking Loss is designed for personalized ranking in implicit feedback, which maximizes the score difference between interacted and non-interacted pairs. 
    {\small\begin{displaymath}
            \mathcal{L}_{\text{BPR}} = -\frac{1}{|D|}\sum_{x\in X}\sum_{y\in\mathcal{I}_x,y^-\in\mathcal{I}_x^-}\log\left(\sigma\left(o_y^{(x)}-o_{y^-}^{(x)}\right)\right)
    \end{displaymath}}
    \item \textbf{BCE}~\citep{he2017neural}: Binary Cross-Entropy Loss regards all interactive pairs as positives and uninteractive pairs as negatives.
    {\small\begin{displaymath}
        \mathcal{L}_{\text{BCE}} = -\sum_{(x,y)\in D}\log\sigma\left(o_y^{(x)}\right)-\sum_{(x,y)\notin D}\log\left(1-\sigma\left(o_{y}^{(x)}\right)\right)
    \end{displaymath}}
    \item \textbf{Sparsemax}~\citep{martins2016softmax}: Sparsemax is a variant of Softmax that returns sparse posterior distributions by assigning zero probability to some classes.
   {\small \begin{displaymath}
        \mathcal{L}_{\text{Sparse}}=-\sum_{(x,y)\in D}\left(-o_y^{(x)}+\frac{1}{2}\sum_{y'\in S{(x)}}^N\left({o_{y'}^{(x)}}^2-{\tau(x)}^2\right)+\frac{1}{2}\right)
    \end{displaymath}}
    where $\tau{(u)} = \left(\left(\sum_{y'\in S{(x)}}o_{y'}^{(x)}\right)-1\right) / {|S{(x)})|}$, 
    
    $S{(x)}=\{y\in\mathcal{I}\mid \operatorname{sparsemax}_y\left(\bm{o}^{(x)}\right)>0\}$ and 
    $\operatorname{sparsemax}\left(\bm{o}^{(x)}\right) = \underset{\bm{p}\in\Delta^{K-1}}{\arg\min}\|\bm{p}-\bm{o}^{(x)}\|^2$.
    \item \textbf{SSM}~\citep{covington2016deep,yi2019sampling} approximates \textbf{SM} Loss through Eq.(\ref{sampled softmax loss}).
    \item \textbf{SM} maximizes the probability of the observed items normalized over all items by Eq.\eqref{rec loss form}.
\end{itemize}
As for recommendation-specific objectives, we have
\begin{itemize}
 \item \textbf{UIB}~\citep{zhuo2022learning}: introduces a learnable auxiliary score $b_x$ for each user to represent the User Interest Boundary and penalizes samples that exceed the decision boundary. With $\phi(\cdot)$ being MarginLoss, the loss form is given as follows:
    {\begin{equation*}
        \mathcal{L}_{\text{UIB}} = \sum_{(x,y)\in D}\phi(b_x-o_y^{(x)})+\sum_{(x,y)\notin D}\phi(o_y^{(x)}-b_x)
    \end{equation*}}
    \item \textbf{SML}~\citep{li2020symmetric} improves the limitation of \textbf{CML} by introducing dynamic margins. Let $d(x,y)$ denote the distance function between embeddings of user $x$ and item $y$, the loss form is given as:
    \begin{small}
    {\begin{gather*}
        \mathcal{L}_{\text{SML}} = \sum_{(x,y)\in D}\sum_{(x,y^-)\notin D}([d(x,y)-d(x,y^-)+m_x]_+\\
        +\lambda[d(x,y)-d(y,y^-)+n_y]_+)-\gamma(\frac{1}{|X|}\sum_x m_x +\frac{1}{|Y|}\sum_y n_y),\\ m_x\in(0,l],n_y\in(0,l],\forall x,y\in X,Y
    \end{gather*}}\end{small}
    \item \textbf{WRMF}~\citep{hu2008collaborative} uses \textbf{ALS} methods to optimize the loss value in Eq.(\ref{eq:WSL}).
    \item \textbf{RG$^2$} and \textbf{RG$^\times$} utilizes \textbf{ALS} in Alg.(\ref{alg:wALS}) to optimize the loss value in Eq.(\ref{eq:RG2 MF form}) and Eq.(\ref{eq:RG2 MF form}). Considering the operability of experiments, the coefficients $W$ and $V$ in equations are replaced by hyperparameters $\alpha, \beta$.
\end{itemize}


\subsection{Implementation Details}\label{appdix.imp}
We conduct all experiments on a highly-modularized library \textbf{RecStudio}(\cite{lian2023recstudio}). The loss functions and baselines are implemented on a \textbf{MF} model with embedding size set to 64 on each layer. For \textbf{BPR} and \textbf{BCE}, we draw 10, 20, 10 and 20 negative samples uniformly for each positive in the training procedure for MovieLens, Electronics, Steam and Wiki, respectively. For \textbf{SSM}, the proposal distribution is set as uniform sampling, and the numbers of negative samples are set as 100, 200, 100 and 1000, respectively. As for \textbf{UIB} and \textbf{SML}, the negative sampling numbers are 10, 10, 1 and 1, 100, 50 for all recommendation datasets. We use a single Nvidia RTX-3090 with 24GB memory in training for all methods. Except for \textbf{WRMF},\textbf{RG$^2$} and \textbf{RG$^\times$}, all baselines are optimized with \textbf{Adam}(\cite{kingma2014adam}), which is a variant of \textbf{SGD}. As for \textbf{SGD} optimization, the learning rate and weight decay are tuned in \{0.1,0.01,0.001\} and \{$0, 10^{-6},10^{-5},10^{-4}$\}, respectively. As for \textbf{WRMF}, the hypermeters $\lambda$ and $\alpha$ are tuned in \{0, 0.1, 0.01, 0.001\} and \{0.5, 1, 2, 4, 8\}, respectively. And for \textbf{RG}s, $\lambda$, $\alpha$ and $\beta$ are tuned in \{0, 1, 0.5, 0.1, 0.05, 0.01, 0.005, 0.001\}. 


\vskip 0.2in

\bibliography{sample}

\end{document}

%% file: section/intro.tex
\section{Introduction}
\label{introduction}
Similarity learning serves as a foundational paradigm in machine learning research, with profound implications across multiple interdisciplinary domains including image retrieval (\cite{chopra2005learning}, \cite{wang2014learning}, \cite{chen2020simple}), natural language processing (\cite{mikolov2013distributed},\cite{reimers2019sentence}, \cite{gao2021simcse}), bioinformatics (\cite{senior2020improved}) and recommender systems(\cite{koren2009matrix}, \cite{rendle2012bpr}, \cite{he2017neural}). A fundamental formulation unifying these applications involves learning pairwise relationships between distinct entity classes - conventionally termed as left (user queries, input modalities) and right (candidate items, targetted concepts) components. The objective resides in developing a parametric mapping function that projects entities from both classes into a shared latent space, where geometric proximity (measured through inner products, distance metrics, etc.) accurately reflects application-specific similarity.

Despite having undergone extensive theoretical refinement and practical implementation through decades of research, this paradigm continues to confront emerging challenges in contemporary applications. \textit{First}, since similarity learning frameworks frequently reduce to fundamental machine learning objectives formulated as regression, classification, or ranking problems. These formulations typically employ task-specific evaluation metrics (as summarized in Table~\ref{tab:metrics}). While regression objectives are typically smooth, classification and ranking tasks generally involve discrete or non-smooth formulations, introducing NP-hard optimization challenges that demand specialized algorithmic solutions. \textit{Second}, with the advent of modern large-scale datasets, the cardinality of instances in both component spaces increasingly exceeds the scalability limits of conventional training frameworks. Both challenges have given rise to the research domain of \textit{Extreme Similarity Learning}.

\begin{table}[htbp]
\centering
\caption{Machine Learning Tasks and Representative Evaluation Metrics}
\label{tab:metrics}
\begin{tabular}{p{4.5cm}p{9.8cm}}
\toprule
\textbf{Task} & \textbf{Metrics} \\
\midrule
Regression & MSE, RMSE, MAE, R$^2$ Score, etc.\\
Classification & Acc, Precision, Recall, F1 Score, AUC-ROC, Log Loss, etc. \\
Ranking &  NDCG, MAP, MRR, etc. \\
\bottomrule
\end{tabular}
\end{table}

To address the first challenge, practical implementations have developed convex relaxations or approximations, namely \textbf{surrogate loss functions}, as a principled approach. Effective surrogates are expected to satisfy several key properties: smoothness for gradient-based optimization (\cite{amari1993backpropagation}); robustness to noise and outliers (\cite{huber1981robust}); and crucially statistical consistency with the target metric (\cite{zhang2004statistical}, \cite{bartlett2006convexity}). (Bayes-)Consistency ensures that surrogate losses share a consistent optimization direction with the targeted objective under abundant training samples, avoiding the optimization from falling into suboptimals. For classification tasks, the consistency properties between losses (logistic, cross-entropy, etc.) and 0-1 risk have been thoroughly studied by \cite{zhang2004statistical}, \cite{tewari2007consistency}. Yet for ranking scenarios, though several related works (\cite{cossock2006subset},\cite{ravikumar2011ndcg}) analysed the relationship between losses and ranking metrics such as Normalized Discounted Cumulative Gain (\textbf{NDCG}), the techniques of analysing or designing Bayes-consistent loss formulations under a unified ranking context remain theoretically underexplored, with many existing approaches relying on heuristic designs rather than formal guarantees. This theoretical gap raises concerns about potential misalignment between optimization objectives and ranking metrics. 

As for the second challenge, \textbf{negative sampling}(\cite{gutmann2010noise},\cite{rendle2012bpr}, \cite{mikolov2013distributed}) has been widely adopted in loss function design for extreme similarity learning to improve training efficiency. This strategy divides surrogate loss functions into two categories: \textbf{sampling} methods and \textbf{non-sampling} methods. Sampling methods typically employ efficient sampling techniques to select negative samples from a large corpus, either randomly or based on specific strategies for training, which significantly reduces computational complexity and accelerates the training process. Conversely, non-sampling methods like \textbf{SM} or \textbf{WSL} (\cite{hu2008collaborative}, \cite{xin2018batch}) utilize all samples as negatives, thereby potentially achieving more accurate ranking performance at the cost of higher complexity. While both methods have demonstrated empirical success in practice, most existing discussions on sampling vs. non-sampling are based on qualitative observations, with limited analysis exploring their theoretical relationships.

At this point, the understanding of surrogate ranking losses faces complex challenges. Fortunately, these challenges can be addressed by leveraging the advantages of \textbf{SM}, which provides a connection between solid theoretical foundation and flexible loss form transformations. As highlighted in \cite{sun2019bert4rec} and \cite{rendle2021item}, \textbf{SM} aims to maximize the likelihood of positive instances over negatives, thereby aligning its objective with ranking metrics such as Discounted Cumulative Gain (\textbf{DCG}) (\cite{bruch2019analysis}, \cite{ravikumar2011ndcg}). This alignment offers substantial benefits for ranking performance. Besides, due to the non-linearity of softmax operation, which complicates the optimization process particularly when applied to large-scale corpora, a sampled version of \textbf{SM}, known as Sampled Softmax (\textbf{SSM}) Loss (\cite{covington2016deep}, \cite{yi2019sampling}), has been proposed. \textbf{SSM} strikes a balance between ranking performance and computational efficiency by selectively limiting the number of negative samples used during training. Based on \textbf{SM} and \textbf{SSM}, the design of Bayes-consistent surrogate ranking loss and the explanation of the intrinsic theoretical relationship between sampling and non-sampling loss functions become possible.

In this paper, we revisit the designs of surrogate ranking loss functions. We start with a thorough discussion of \textbf{SM} and \textbf{SSM}, and then introduce novel surrogate squared-form non-sampling loss, namely \textbf{R}anking-\textbf{G}eneralizable \textbf{squared}  loss and \textbf{R}anking-\textbf{G}eneralizable \textbf{interactive}  loss (\textbf{RG$^2$} and \textbf{RG$^\times$}) , drawn from Taylor approximations of \textbf{SM} and \textbf{SSM}. The Bayes-consistency of \textbf{RG} losses with \textbf{DCG}, inherited from \textbf{SM}, is well-established through rigorous analysis. As a typical application of the \textbf{RG} losses, we choose a ranking similarity learning task with matrix factorization (\textbf{MF}) backbone, which integrates the \textbf{ALS} optimization method, and derive generalization upper bounds based on standard generalization theories. This integration enhances ranking performance compared to the existing approaches, as the \textbf{RG} losses aligns closely with the ranking metric and benefits from the rapid convergence rate of \textbf{ALS} over stochastic gradient descent (\textbf{SGD}).

The main contributions of this article are as follows: 
\begin{itemize}[leftmargin=*] 
\item We establish a new framework on analysing and designing ranking objective-consistent surrogate loss functions inspired by Taylor expansions and propose novel squared loss functions, \textbf{RG$^2$} and \textbf{RG$^\times$}, which serve as efficient approximations of \textbf{SM/SSM} Loss. This approach uncovers the connection between squared loss and ranking metrics, providing new insights into techniques of designing surrogate losses. 
\item We propose an understanding on the inherent relationship between sampling and non-sampling methods. Derived from \textbf{SSM}, \textbf{RG} losses are non-sampling methods whose coefficients of positive and negative interactions are inherently determined by the sampling size.
\item We incorporate \textbf{RG}s into \textbf{MF} backbone and provide generalization upper bounds. Additionally, we integrates the \textbf{ALS} optimization method and analyze the convergence rate to show the superiority of squared-form loss.  
\item Experimental results on real-world datasets concerning recommender system and link prediction demonstrate the effectiveness of \textbf{RG}s, showcasing significant efficiency improvements without compromising performance of extreme similarity learning.
\end{itemize}

The remainder of this paper is organized as follows: Section 2 introduces related works of consistency property and typical surrogate loss designs of extreme similarity learning. Section 3 presents foundational concepts and methodologies relevent to all discussed loss functions. Section 4 carefully derives the \textbf{RG} framework and proposes all theoretical results. Experimental results and discussions follow in Section 5, with conclusion in Section 6.

%% file: section/related.tex
\section{Related Works}

\subsection{Consistency}
Bayes-consistency is a fundamental concept in understanding the asymptotic behavior of surrogate losses, particularly in classification and ranking tasks. For classification, extensive theoretical results \citep{zhang2004statistical, bartlett2006convexity,tewari2007consistency} demonstrate that losses such as logistic loss and hinge loss are consistent with the 0-1 risk under specific conditions. \cite{wydmuch2018no} and \cite{menon2019multilabel} modeled recommendation tasks as multi-label classification problems and analyzed the Bayes-consistency of \textbf{SM} Loss and One-vs-All(\textbf{OvA}) Loss with respect to metrics like Precision@k and Recall@k. These studies emphasize the importance of convexity and margin-based properties in ensuring consistency, providing a solid foundation for designing surrogate losses in classifications.

In ranking tasks, the study of Bayes-consistency is more complex due to the structured nature of ranking metrics such as \textbf{NDCG} and \textbf{MAP}. Early work by \cite{cossock2006subset} and \cite{ravikumar2011ndcg} explored the relationship between surrogate losses and ranking metrics, showing that certain convex surrogates can achieve consistency with ranking objectives. For example, \cite{ravikumar2011ndcg} demonstrated that surrogate losses aligned with \textbf{NDCG} can achieve Bayes-consistency under specific conditions. However, these studies primarily focused on pairwise and listwise ranking approaches, leaving a gap in understanding consistency for more specialized ranking scenarios.

Besides, since Bayes-consistency is an asymptotic property, its practical implications can be limited in finite-sample settings. This has led to the exploration of alternative notions of consistency, such as $\mathcal{H}$-consistency, which takes the hypothesis set $\mathcal{H}$ into account. Introduced by \cite{long2013consistency}, $\mathcal{H}$-consistency provides a more nuanced understanding of learning consistency by examining the relationship between surrogate losses and hypothesis sets. Recent advances in $\mathcal{H}$-consistency theory, such as those by \cite{awasthi2022h, awasthi2022multi}, have provided stronger guarantees by deriving bounds on the target loss estimation error in terms of the surrogate loss estimation error. These results are particularly valuable in binary and multi-class classification, as they bridge the gap between asymptotic properties and finite-sample performance. Nevertheless, the focus on Bayes-consistency remains critical, as it provides the theoretical foundation for understanding the alignment between surrogate losses and target metrics, guiding the design of effective learning algorithms in both classification and ranking tasks.

\subsection{Surrogate Loss Functions for Ranking}

The choice of surrogate loss functions plays a crucial role in capturing the similarities between objects to be trained. Based on whether sampling is involved, surrogate losses can be categorized into non-sampling and sampling methods.

\subsubsection{Non-sampling Loss}
Non-sampling losses directly compute the similarity or dissimilarity between samples or sample pairs without requiring explicit sampling strategies.

For instance, \textbf{squared loss}, which directly measures the difference between predicted and true values, not only possesses a regression nature but also shows demonstrated ranking performance \citep{rendle2021item, chen2023revisiting, chen2020efficient}. Squared loss benefits from the absence of non-linear operations, enabling closed-form solutions and efficient optimization using methods like \textbf{ALS} \citep{hu2008collaborative,takacs2012alternating} and coordinate descent \citep{bayer2017generic}. However, the limited correlation between squared loss and ranking metrics remains an area of ongoing research.

\textbf{SM loss}, closely related to cross-entropy loss, is widely used in classification tasks. Unlike squared loss, which treats ranking as a regression task, \textbf{SM} models user interests as a multinomial distribution \citep{shen2014learning,liang2018variational,sun2019bert4rec} and aims to maximize the likelihood function. \textbf{SM} aligns well with ranking metrics \citep{bruch2019analysis,ravikumar2011ndcg,huang2023cooperative}, making it particularly effective for implicit feedback scenarios. However, the exponential operation in \textbf{SM} precludes closed-form solutions or higher-order gradients over all objects, necessitating optimization via \textbf{SGD}. To address this, researchers have proposed approximations such as \textbf{SSM}\citep{wu2022effectiveness} and sparse approximations of the probability vector \citep{martins2016softmax}.

\subsubsection{Sampling Loss}
While non-sampling methods that consider relationships across the entire corpus generally achieve superior accuracy, their computational costs can be prohibitive, especially for large-scale datasets. Negative sampling has emerged as an efficient alternative, reducing computational overhead by sampling a subset of negative instances. Started from direct uniform sampling strategies, prominent examples include \cite{burges2010ranknet}, \cite{rendle2012bpr}, \cite{mikolov2013distributed}, and \cite{Schroff2015}, which uniformly samples negative instances to distinguish positive ones. To improve sample quality, strategies aiming at mining hard negative samples have been proposed, including \cite{weston2011wsabie}, \cite{zhang2013optimizing}, \cite{rendle2014improving}. Another widely used approach is importance sampling, which estimates gradient expectations like the operations in \textbf{SSM} \citep{yi2019sampling,bengio2008adaptive}. While these methods effectively reduce computational costs, they introduce biases and variances in gradient estimation due to limited sample sizes and biased sampling distributions. These challenges are exacerbated when approximating higher-order gradients, rendering advanced optimization algorithms inapplicable.

%% file: section/preliminary.tex
\section{Preliminaries}

This section introduces foundational concepts and methodologies relevant to the loss functions commonly used in ranking tasks. Specifically, we examine the framework of \textbf{WSL}, \textbf{SM}, and \textbf{SSM}, with a detailed description of consistency properties.



\subsection{Ranking Similarity Learning Framework}

Consider a ranking scenario where triplets $(x, y, y^-)$ are drawn from a joint distribution $\mathcal{P}$ defined over $X \times Y \times Y$, where $x\in X$ represents a context, $y\in Y$ a positive object, and $y^-\in Y$ a negative object. We denote $M = |X|$ and $N = |Y|$ for simplicity. A model $f: X \times Y \to \mathbb{R}$ predicts a similarity score, indicating whether object $y$ is more similar to context $x$ than object $y^-$, i.e., $f(x, y) > f(x, y^-)$. In large-scale ranking scenarios, ranking objectives are often reduced to classification-like formulations due to sparse positive interactions, where only few interactions are observed as positive among a large candidate set. Let $\mathcal{D}$ denote the data distribution over pairs $(x, y)$, where $y$ is observed as a positive interaction under context $x$. A set of observed positive interactions $D = {(x_k, y_k)}_{k=1}^{|D|}$ is assumed to be drawn i.i.d. from $\mathcal{D}$. Let $o_y^{(x)}=f_\theta(x, y)$ represent the output of model $f$, denoting the score of the predicted similarity score for object $y$ in context $x$.

To measure the ranking quality of different algorithms, representative ranking metrics include \textbf{DCG} (with its normalized version \textbf{NDCG}), \textbf{MAP}, and \textbf{MRR}, whose definitions are given as follows:

\begin{definition}\label{def:DCG&NDCG}
\begin{gather}
    \operatorname{DCG}(\bm{o}^{(x)},D) = \sum_{y \in Y} \frac{2^{r_{x,y}} - 1}{\log(1 + \pi^{(x)}(y))}, \\
    \operatorname{NDCG}(\bm{o}^{(x)},D) = \frac{\operatorname{DCG}(\bm{o}^{(x)},D)}{\max_{\bm{o}} \operatorname{DCG}(\bm{o}, D)},
\end{gather}
where $\bm{o}^{(x)} \in \mathbb{R}^N$ is the score vector for instance $x$, $\pi^{(x)}(y)$ denotes the rank of item $y$ under $\bm{o}^{(x)}$, and $r_{x,y} \in \{0,1\}$ indicates whether item $y$ is relevant to $x$ (i.e., whether $(x, y) \in D$).
\end{definition}
As for \textbf{MRR} and \textbf{MAP},
\begin{definition}\label{def:MRR&MAP}
\begin{gather*}
    \operatorname{MRR} = \frac{1}{|X|} \sum_{x=1}^{M} \sum_{i=1}^{N} \frac{\operatorname{rel}_i^{(x)}}{i} \cdot \prod_{j=1}^{i-1} (1 - \operatorname{rel}_j^{(x)}), \\
    \operatorname{MAP} = \frac{1}{|X|} \sum_{x=1}^{M} \frac{1}{|\mathcal{I}_x|} \sum_{i=1}^{N} \left( \frac{\sum_{m=1}^{i} \operatorname{rel}_m^{(x)}}{i} \cdot \operatorname{rel}_i^{(x)} \right),
\end{gather*}

where $\operatorname{rel}_i^{(x)} \in \{0, 1\}$ indicates whether the item at rank $i$ is relevant to context $x$. $\mathcal{I}_x$ denotes the set of relevant items for context $x$.

\end{definition}

In practice, directly optimizing ranking metrics above is often intractable due to their non-differentiability. Instead, surrogate loss functions $\ell : \mathbb{R} \to \mathbb{R}^+$ are introduced to approximate the ranking objective. Common surrogate losses take the following form:
\begin{equation}
\begin{aligned}
    &\mathcal{L}_{\operatorname{pairwise}}(\theta)=\mathbb{E}_{(x,y,y^-)\sim\mathcal{P}}\left[\ell(o_y^{(x)}-o_{y^-}^{(x)})\right]\\
    &\mathcal{L}_{\operatorname{pointwise}}(\theta)=\mathbb{E}_{(x,y)\sim\mathcal{D}}\left[\ell(o_y^{(x)})\right]
\end{aligned}    
\end{equation}
while we focus on point-wise losses (\textbf{SM}, \textbf{SSM}, \textbf{WSL}, etc.) in subsequent contents.

\subsection{Softmax Loss}

\textbf{SM} Loss was first introduced to modeling the probability distribution of multi-class classifications and was soon generalized to various machine learning tasks and applications. The softmax operation transforms the logits into:
\begin{equation}\label{eq:softmax function}
p_{\text{SM}}(o_y^{(x)}) = \frac{\exp(o_y^{(x)})}{\sum_{y'\in Y}\exp(o_{y'}^{(x)})}
\end{equation}
which essentially converts the output vector $\bm{o}^{(x)}=[o_{y_1}^{(x)},\cdots, o_{y_N}^{(x)}]$ into a probability distribution for each input context $u$. Subsequently, \textbf{SM} Loss, also commonly referred to as \textbf{Categorical Cross-Entropy} (\textbf{CCE}), is computed for each context-object pair $(x, y)\sim\mathcal{D}$ as follows:
\begin{equation}
\label{single loss form}
\ell(o^{(x)}_y) = -\log\left(p_{\text{SM}}(o_{y}^{(x)})\right).
\end{equation}
The standard form of \textbf{SM} can be expressed as:
\begin{equation}
\label{rec loss form}
\mathcal{L}_{\text{SM}} = \mathbb{E}_{(x,y)\sim\mathcal{D}}\left[-\log\left(p_{\text{SM}}(o_{y}^{(x)})\right)\right] +\lambda\psi(\theta)
\end{equation}
where $\lambda\psi(\theta)$ stands for some regularization term (usually $l_2$-norm) in practice.

The widespread application of \textbf{SM} is not only attributed to its excellent ranking performance but also to its profound theoretical properties. The efficacy of \textbf{SM} can be understood from two perspectives. Firstly, \textbf{SM} indirectly regulates the lower bound of \textbf{NDCG}, i.e.,

\begin{proposition}(\cite{bruch2019analysis})
    Softmax loss is a bound on mean Normalized Discounted Cumulative Gain in log-scale, i.e.
    \begin{equation}
        \mathbb{E}_{\mathcal{D}}\left[\log \operatorname{NDCG}\right]\geq \mathbb{E}_{\mathcal{D}}\left[\log\left(p_{\operatorname{SM}}(o_{y}^{(x)})\right)\right]
    \end{equation}
\end{proposition}

This characteristic makes it particularly effective in handling ranking-oriented tasks. Secondly, from a more fundamental viewpoint, \textbf{SM} is both \textbf{Top-$k$ calibrated} and \textbf{DCG-consistent}, providing a more intrinsic guarantee of its effectiveness in both classification and ranking tasks. (See more discussion in Appendix.\ref{Appdix.SMConsistent})

\begin{definition}[Bregman Divergence]
Given a convex, differentiable function $\phi: \mathbb{R}^N \to \mathbb{R}$, the Bregman divergence between $\bm{s}, \bm{t} \in \mathbb{R}^N$ is defined as
\[
D_\phi(\bm{s}, \bm{t}) = \phi(\bm{t}) - \phi(\bm{s}) - \langle \nabla \phi(\bm{s}), \bm{t} - \bm{s} \rangle.
\]
\end{definition}

\begin{proposition}[\cite{yang2020consistency}]
The Softmax Loss can be rewritten as a Bregman divergence:
\[
-\log(p(o_y^{(x)})) = D_{\phi}(g(\bm{o}^{(x)}), \bm{e}_y),
\]
where $\phi(\bm{o}) = \sum_y o_y \log o_y$, and $g(\bm{o})_y = p(o_y)$ denotes the softmax mapping. Since $\phi$ is strictly convex and differentiable, and $g$ is inverse top-$k$ preserving, the Softmax Loss is top-$k$ calibrated.
\end{proposition}

\begin{proposition}[\cite{ravikumar2011ndcg}]\label{prop.SMConsist}
    Following the form of Bregman divergence above. Since $\phi$ is strictly convex and differentiable, and $g$ is inverse order-preserving, the softmax loss is thereby $\operatorname{DCG}$-consistent.
\end{proposition}

This reveals \textbf{SM}'s theoretical soundness in optimizing ranking-oriented objectives, offering theoretical guarantees of applying \textbf{SM} into ranking scenarios.

\subsection{Sampled Softmax}

The standard softmax function, as described in Eq.(\ref{eq:softmax function}), suffers from a significant computational bottleneck due to the partition function $Z = \sum_{y\in Y}\exp(o_{y}^{(x)})$ over the entire candidate set $Y$, which can be prohibitively expensive when $N$ is overwhelmingly large. To address this issue, \textbf{SSM} was introduced as an efficient approximation. Instead of computing over all objects, \textbf{SSM} samples a subset of $n$ instances from a pre-defined distribution $\mathcal{D}'$, namely $[n]\sim\mathcal{D}'$, and computes the softmax function over this subset:
\begin{equation}
p_{\text{SSM}}(o_y^{(x)}) = \frac{\exp(o_y^{(x)})}{\exp(o_y^{(x)})+\sum_{y'\in[n]}\exp(o_{y'}^{(x)})}.
\end{equation}
Assume that each negative instance is sampled with probability $q_y$ with
replacement. As mentioned in \cite{bengio2008adaptive}, the following correlation of logits in \textbf{SSM} leads to an unbiased approximation of \textbf{SM},
\begin{equation}
    \tilde{o}_y^{(x)}=\begin{cases}
        o_y^{(x)}-\log(Nq_y) & \text{if}\ (x,y)\notin\mathcal{D}\\
        o_y^{(x)} & \text{elsewise}
    \end{cases}
\end{equation}
It is worth noting that uniform sampling does not require any adjustments with $q_y=\frac{1}{N}$. The loss function for \textbf{SSM} can then be written as:
\begin{equation}
\label{sampled softmax loss}
\mathcal{L}_{\text{SSM}} = \mathbb{E}_{(x,y)\sim\mathcal{D},[n]\sim\mathcal{D}'}\left[-\log\left(p_{\operatorname{SSM}}(\tilde{o}_{y}^{(x)})\right) \right]+\lambda\psi(\theta).
\end{equation}
It significantly reduces the computational complexity of the softmax operation, making it feasible for large-scale ranking tasks where $N$ is extremely large.

\subsection{Weighted Squared Loss}
\textbf{WSL} is a typical non-sampling loss for handling similarity learning tasks. It assigns different weights to postive and negative instances, indicating greater importance in positive compared to negative ones. A specific form of weighted squared loss follows:
\begin{equation}\label{eq:WSL}
\mathcal{L}_{\text{WSL}} = \mathbb{E}_{(x,y)\in X\times Y}\left[w_{x,y}\left(o_y^{(x)}-r_{x,y}\right)^2\right]+ \lambda\psi(\theta)
\end{equation}
where
\begin{equation*}
r_{x,y}= \left\{
\begin{split}
1,\quad & (x,y)\in D\\
0, \quad& (x,y)\notin D
\end{split}\right.\quad
\end{equation*}
is a common binary labeling setup, indicating whether the interaction $(x,y)$ is observed, and $w_{x,y}$ is determined according to heuristic design choices, reflecting the relative importance or confidence associated with different samples. A straightforward example is the \textbf{WRMF} loss (\cite{hu2008collaborative}) in implicit feedback-based recommender systems, where $w_{x,y}$ is set to $\alpha+1$ for positive instances and $1$ for negative ones. Other examples include AllVec\citep{xin2018batch} for word embeddings and TransRW\citep{mai2018support} for knowledge graph embeddings, whose $w_{x,y}$ are determined based on word co-occurrence frequencies and structural measures, respectively.

\textbf{WSL} is frequently employed in dual encoders. iALS (\cite{rendle2022revisiting}) utilizes this loss to optimize matrix factorization models, while SAGram (\cite{krichene2018efficient}) applies it to optimize non-linear encoders. A noteworthy advantage of \textbf{WSL} stems from the applicability of \textbf{ALS} or higher-order gradient descent optimization methods like Newton's method, which involve updating the left and right latent matrices using the closed-form solutions or the second-order Hessian matrix. The competitive performance demonstrated by iALS further emphasizes the efficacy of the weighted squared loss. However, the optimization objectives associated with this loss currently lack a clear theoretical foundation to fully understand the underlying mechanism, as the squared loss formulation does not explicitly optimize ranking-based objectives. Consequently, further investigation is necessary to uncover the underlying principles.


%% file: section/method.tex
\section{Theoretical Results}

This section presents the following theoretical developments. First, by applying Taylor expansions to the \textbf{SM} and \textbf{SSM} loss functions, we derive approximations of their gradient optimization directions and introduce two novel loss formulations, referred to as \textbf{RG$^2$} and \textbf{RG$^\times$}. Among these, \textbf{RG$^2$} adopts a squared-form loss structure which, in contrast to the heuristic design of \textbf{WSL}, exhibits a closer alignment with the gradient direction of \textbf{SM}. Furthermore, we provide formal proof that the proposed \textbf{RG} losses are Bayes-consistent with respect to the ranking metric \textbf{DCG}, sharing the same consistency property as \textbf{SM}.

To further leverage the advantages of the \textbf{RG} losses, we incorporate them into \textbf{MF}-based backbones and establish corresponding generalization guarantees. In addition, we exploit the structural properties of the \textbf{RG} losses to apply the highly efficient \textbf{ALS} optimization, thereby significantly enhancing training efficiency.



\subsection{Efficient Approximation by Taylor Expansion}
Reflecting on the loss form in Eq.(\ref{single loss form}), we start with a direct Taylor expansion at any $\bm{o}_0\in\mathbb{R}^{N}$ with respect to any given $\bm{o}^{(x)}\in\mathbb{R}^{N}$:
\begin{equation}\label{eq.TaylorAny}
    \ell(\bm{o}^{(x)})\overset{\operatorname{Taylor}} = \ell(\bm{o}_0)+{\nabla\ell(\bm{o}_0)}^\top\bm{o}^{(x)}+\frac{1}{2} {\bm{o}^{(x)}}^\top\nabla^2\ell(\bm{o}_0)\bm{o}^{(x)} + o(\|\bm{o}^{(x)}-\bm{o}_0\|^2)
\end{equation}
where 
\begin{align*}
    \text{0-th}:\quad&\ell(\bm{o}_0)=-\log(p(\bm{o}_0))\\
    \text{1-st}:\quad&\nabla{\ell}(\bm{o}_0) = [\frac{\partial\ell}{\partial o_{y_1}},\cdots,\frac{\partial\ell}{\partial o_{y_N}}]^\top=[p(\bm{o}_0)_1,p(\bm{o}_0)_2,\cdots,p(o_{0})_i-1,\cdots,p(o_{0})_N]^\top\\
    \text{2-nd}:\quad&\frac{\partial^2\ell}{\partial (o_y^{(x)})^2} = p(o_y^{(x)})(1-p(o_y^{(x)})),\frac{\partial^2\ell}{\partial (o_y^{(x)})\partial(o_{y'}^{(x)})} = -p(o_y^{(x)})p(o_{y'}^{(x)})\\
    &\implies [\nabla^2\ell(\bm{o}_0)]_{y,y'}= p(\bm{o}_0)_y\left(\delta_{y,y'}-p(\bm{o}_0)_{y'}\right)
\end{align*}

Arguably, Eq. (\ref{eq.TaylorAny}) is a good approximation of \textbf{SM} at first and second-order gradients. However, directly obtaining such a result is not enough to solve the problem of nonlinear computation of \textbf{SM}, since the gradients contains these operations as well. To address this issue, we take a simplified yet effective approximation by expanding at $\bm{o}_0 = \bm{0}$ and neglecting higher-order terms, i.e.,
\begin{equation}\label{eq.Taylor0}
\begin{aligned}
     \ell(\bm{o}^{(x)})&\overset{\operatorname{Taylor}}{=}\ell(\bm{0})+{\nabla\ell(\bm{0})}^\top\bm{o}^{(x)}+\frac{1}{2} {\bm{o}^{(x)}}^\top\nabla^2\ell(\bm{0})\bm{o}^{(x)}\\
     &=\log N - o_{i}^{(x)}+\frac{1}{N}\bm{1}_N^\top\bm{o}^{(x)}+ \frac{1}{2N} {\bm{o}^{(x)}} ^\top(\bm{I}_N-\frac{1}{N}\bm{1}_N\bm{1}_N^\top)\bm{o}^{(x)}
\end{aligned}
\end{equation}

The choice of $\bm{o}_0=\bm{0}$ is based on the following two reasons: 1. The structure of the gradient information at the $\bm{0}$-vector is very simple, which greatly simplifies the computation of the loss function. 2. The optimal solution is typically constrained close to the zero vector due to regularization. Based on the form in Eq. (\ref{eq.Taylor0}), we have

\begin{equation}
\begin{aligned}
        \ell(\bm{o}^{(x)}) &= \log N - o_{y}^{(x)} + \frac{1}{N} \bm{1}_N^\top \bm{o}^{(x)} + \frac{1}{2N} {\bm{o}^{(x)}} ^\top \bm{o}^{(x)}-\frac{1}{2N^2}(\bm{1}_N^\top\bm{o}^{(x)})^2\\
    &=\left(\log N-\frac{1}{2}\right) - o^{(x)}_{y}+\frac{1}{2N}\|\bm{o}^{(x)}+\bm{1}_N\|^2-\frac{1}{2N^2}(\bm{1}_N^\top\bm{o}^{(x)})^2\\
    &\propto- o^{(x)}_{y}+\frac{1}{2N}\|\bm{o}^{(x)}+\bm{1}_N\|^2-\frac{1}{2N^2}(\bm{1}_N^\top\bm{o}^{(x)})^2
\end{aligned}
\end{equation}

This leads to two possible treatment strategies: 1. based on the assumption of independence between samples, we ignore the interaction term between the scores of different objects, as we did in \cite{pu2024learning}. 2. we take all parts into consideration, as shown as follow:
\begin{equation}\label{eq.RG}
\begin{aligned}
    \mathcal{L}_{\operatorname{RG^2}}&=\mathbb{E}_{\mathcal{D}}\left[- o^{(x)}_{y}+\frac{1}{2N}\|\bm{o}^{(x)}+\bm{1}_N\|^2\right]+\lambda\psi(\theta)\\
\mathcal{L}_{\operatorname{RG^\times}}&=\mathbb{E}_{\mathcal{D}}\left[- o^{(x)}_{y}+\frac{1}{2N}\|\bm{o}^{(x)}+\bm{1}_N\|^2-\frac{1}{2N^2}(\bm{1}_N^\top\bm{o}^{(x)})^2\right]+\lambda\psi(\theta)
\end{aligned}
\end{equation}

An interesting observation is that the realization of such losses can be presented into squared forms by some simple transformations. Starting from $\mathcal{L}_{\text{RG$^2$}}$, we are able to obtain an equivalent formulation as (details in Appendix.~\ref{Appdix.transformation}), 
\begin{equation}
\begin{aligned}\label{eq.RGsquaredform}
    \mathcal{L}_{\operatorname{RG^2}} &=\mathbb{E}_{\mathcal{D}}\left[-o^{(x)}_{y}+\frac{1}{2N}\|\bm{o}^{(x)}+\bm{1}_N\|^2\right]+\lambda\psi(\theta)\\
    &=\frac{1}{|D|}\sum_{(x,y)\in D}\left(-o^{(x)}_{y}+\frac{1} {2N}\|\bm{o}^{(x)}+\bm{1}_N\|^2\right)+\lambda\psi(\theta)\quad (\operatorname{Realization})\\
    &\propto \frac{1}{|D|} \sum_{x\in X,y\in Y}|\mathcal{I}_x| \left(o_y^{(x)}+1 - r_{x,y}\frac{N} {|\mathcal{I}_x|} \right)^2 + \lambda\psi(\theta)
\end{aligned}
\end{equation}
\begin{equation}\label{eq.RG*squaredform}
    \mathcal{L}_{\operatorname{RG^\times}}\propto \frac{1}{|D|} \sum_{x\in X,y\in Y}  |\mathcal{I}_x|\left(\left(o_y^{(x)}+1 - r_{x,y}\frac{N} {|\mathcal{I}_x|} \right)^2-\frac{1}{N}(\bm{1}_N^\top\bm{o}^{(x)})^2\right) + \lambda\psi(\theta)
\end{equation}
where $r_{x,y}$ shares the same definition in Eq.(\ref{eq:WSL}), $|\mathcal{I}_x|$ denotes the number of positive interactions by $x$. We refer to the loss form in Eq.(\ref{eq.RGsquaredform}) a \textbf{RG-squared} (\textbf{RG$^2$}) loss due to its squared form and Eq.(\ref{eq.RG*squaredform}) a \textbf{RG-interactive} (\textbf{RG$^\times$}) loss since it takes all interactions into account. It should be noted that while \textbf{RG$^\times$} appears to approximate \textbf{SM} more tightly, \textbf{RG$^2$} has a simpler form and is able to control the upper bound of \textbf{SM} under a simple assumption, whose proof can be found in Appendix.~\ref{Appdix.UpperBound}. 

\begin{proposition}\label{prop.upperbound}
    Suppose that an extreme similarity learning task satisfies a single-click behavior assumption, then $\mathcal{L}_{\operatorname{SM}}$ is upper bounded by $\mathcal{L}_{\operatorname{RG^2}}$ as they converge to their optimum.
\end{proposition}

Similarly, we perform a similar transformation to \textbf{SSM} as follows:
\begin{displaymath}
\begin{aligned}
    &\tilde{\ell}(\bm{o}^{(x)})\overset{\operatorname{Taylor}}{=}\tilde{\ell}(\bm{0})+{\nabla\tilde{\ell}(\bm{0})}^\top\bm{o}^{(x)}_{[n]}+\frac{1}{2} {\bm{o}^{(x)}_{[n]}}^\top\nabla^2\tilde{\ell}(\bm{0})\bm{o}^{(x)}_{[n]}\\
    =&\log (n+1) - o_{y}^{(x)}+\frac{1}{n+1}\bm{1}_{n+1}^\top\bm{o}^{(x)}_{[n]}+ \frac{1}{2n} {\bm{o}^{(x)}_{[n]}}^\top(\bm{I}_{n+1}-\frac{1}{n+1}\bm{1}_{n+1}\bm{1}_{n+1}^\top)\bm{o}^{(x)}_{[n]}\\
    \propto& -o_y^{(x)}+\frac{1}{2n+2}\|\bm{o}^{(x)}_{[n]}+\bm{1}_{n+1}\|^2-\frac{1}{2(n+1)^2}(\bm{1}_{n+1}^\top\bm{o}^{(x)}_n)^2
\end{aligned}\end{displaymath}
where $\bm{o}^{(x)}_{[n]}=[o_y^{(x)},[o_{y'}^{(x)}]_{y'\in [n]}]$ has 1 positive element and $n$ negative elements. Similar to the transformation in Eq.(\ref{eq.RGsquaredform}), we give out the loss formulations under a uniform sampling distribution where $[n]\sim\mathcal{D}'=\operatorname{Unif}[Y]$,
\begin{equation}
\begin{aligned}\label{eq.RG+}
    \tilde{\mathcal{L}}_{\operatorname{RG^2}} &=\mathbb{E}_{\mathcal{D},\mathcal{D}'}\left[-o^{(x)}_{y}+\frac{1}{2n+2}\|\bm{o}^{(x)}_{[n]}+\bm{1}_{n+1}\|^2\right]+\lambda\psi(\theta)\\
    &=\mathbb{E}_{\mathcal{D}}\left[\mathbb{E}_{\mathcal{D}'}\left[-o^{(x)}_{y}+\frac{1}{2n+2}\|\bm{o}^{(x)}_{[n]}+\bm{1}_{n+1}\|^2|\mathcal{D}\right]\right]+\lambda\psi(\theta)\\
    &\propto \frac{1}{|D|} \sum_{x\in X,y\in Y} w_{x,y} \left(o_y^{(x)}+1 - r_{x,y}\frac{N} {|\mathcal{I}_x|} \right)^2 + \lambda\psi(\theta)\\
    \tilde{\mathcal{L}}_{\operatorname{RG^\times}}&\propto \frac{1}{|D|} \sum_{x\in X,y\in Y}  w_{x,y}\left(\left(o_y^{(x)}+1 - r_{x,y}\frac{N} {|\mathcal{I}_x|} \right)^2-\frac{n+1}{N^2}(\bm{1}_N^\top\bm{o}^{(x)})^2\right) + \lambda\psi(\theta)
\end{aligned}
\end{equation}
where
\begin{equation}
    w_{x,y}=\begin{cases}
        |\mathcal{I}_x| & (x,y)\in D\\
        |\mathcal{I}_x|\cdot\frac{n+1}{N} & (x,y)\notin D
    \end{cases}
\end{equation}

Note that the form of $w_{x,y}$ is very close to the design of allocating different confidence weights to positive and negative instances in \textbf{WSL}. Hence we reveal a theoretical relation between \textbf{SSM} and \textbf{WSL}. To address this point, we would like to have a more in-depth discussion. First, both \textbf{WSL} and our proposed \textbf{RG} losses can be viewed as approximations of \textbf{SM}, with the goal aiming to learn the similarity superiority of positive samples compared to negative samples. Unlike the heuristic design of \textbf{WSL}, our proposed method benefits from the Taylor expansion of \textbf{SM}, allowing the corresponding gradient information to be more accurate, whereas the design of \textbf{WSL} with respect to weights and observations lacks rigorous theoretical guarantees. In fact, the design of the weights $w_{x,y}$ and observations $r_{x,y}$ used in the \textbf{RG} losses, although similar to those in \textbf{WSL}, is fundamentally different in values.

Moreover, our discussion also reveals an inherent connection between sampling and non-sampling losses. Similar operations taken for both non-sampling loss \textbf{SM} and sampling loss \textbf{SSM} yield non-sampling \textbf{RG} losses, and the difference corresponds exactly to the weight design, which suggests that the intrinsic mechanism of adjusting the number of samples is in fact to adjust the weights of the positive and negative samples, which provides a new corroboration of the validity of the sampling loss.

\subsection{Consistency}
\label{consistency section}

Having derived a novel form of the loss function, we now establish its consistency with respect to the target ranking metric. We focus on the \textbf{DCG} metric as defined in Def.~\ref{def:DCG&NDCG}, whose expected risk is given by:
\begin{equation}
    \mathcal{L}_{\operatorname{DCG}}(\bm{o}^{(x)}) = \mathbb{E}_{\mathcal{D}}\left[-\operatorname{DCG}(\bm{o}^{(x)}, D)\right].
\end{equation}
For any surrogate loss function $\phi$, we define its expected risk as:
\begin{equation}
    \Phi(\bm{o}^{(x)}) = \mathbb{E}_{\mathcal{D}}\left[\phi(\bm{o}^{(x)}, D)\right].
\end{equation}
The formal definition of consistency is stated as follows:

\begin{definition}[Bayes-consistency for DCG]
    A surrogate loss $\phi$ is said to be (Bayes-)consistent with respect to $\operatorname{DCG}$ if, for any data distribution $\mathcal{D}$ over $U \times I$ and any sequence $\{\bm{o}^{(x)}_n\}_{n=1}^\infty$, the following holds:
    \begin{equation}
        \Phi(\bm{o}^{(x)}_n) \to \Phi^* \quad \text{implies} \quad \mathcal{L}_{\operatorname{DCG}}(\bm{o}^{(x)}_n) \to \mathcal{L}_{\operatorname{DCG}}^*,
    \end{equation}
    where
    \begin{equation}
        \Phi^* = \min_{\bm{o}^{(x)}} \Phi(\bm{o}^{(x)}), \quad \mathcal{L}_{\operatorname{DCG}}^* = \min_{\bm{o}^{(x)}} \mathcal{L}_{\operatorname{DCG}}(\bm{o}^{(x)}),
    \end{equation}
    assuming the minima are achievable.
\end{definition}

Following this definition, we proceed to prove the \textbf{DCG-consistency} of the proposed $\mathcal{L}_{\operatorname{RG^2}}$. Inspired by \cite{cossock2006subset}, we leverage an auxiliary squared loss as a stepping stone for establishing the consistency of our surrogate.

For simplicity, we define $c_y = \frac{1}{\log(1+y)}$ and $f(x, y) = o_y^{(x)}$. Let $\bm{\pi} = (\pi_1, \dots, \pi_N)$ denote the permutation of object indices, where $\pi_y$ replaces the notation $\pi^{(x)}(y)$ with some fixed context $x$. Under the 0-1 interaction setting, the \textbf{DCG} metric simplifies to:
\begin{equation}
    \operatorname{DCG}(\bm{\pi}, D) = \sum_{y=1}^N c_{\pi_y} r_{x,y}.
\end{equation}

The Bayes-optimal scoring function is defined as:
\begin{equation}
    f_B(x, y) = \mathbb{E}_{\mathcal{D}}[r_{x,y} \mid x],
\end{equation}
which minimizes the expected DCG loss and is assumed to lie within the hypothesis space $\mathcal{H}$. The permutation induced by sorting $f_B(x, y)$ is denoted as $\bm{\pi}^* = (\pi_1^*, \dots, \pi_N^*)$.

We now state the following lemma (proof provided in Appendix~\ref{pf.lem6}):

\begin{lemma}\label{lem.dcgconsist}
    The suboptimality gap of $\operatorname{DCG}$ can be upper-bounded as:
    \begin{equation}
        \mathcal{L}_{\operatorname{DCG}}(\bm{\pi}) - \mathcal{L}_{\operatorname{DCG}}^* \leq \left(2 \sum_{y=1}^N c_y^2 \right)^{\frac{1}{2}} \left( \sum_{y=1}^N \left( f(x, y) - f_B(x, y) \right)^2 \right)^{\frac{1}{2}}.
    \end{equation}
\end{lemma}

As a direct application, consider the squared-form surrogate loss $\phi_B$, defined as:
\begin{gather*}
    \phi_{B}(f, D) = \sum_{y=1}^N \left( f(x, y) - r_{x,y} \right)^2
    = \sum_{y \mid (x, y) \in \mathcal{D}} \left( f(x, y) - 1 \right)^2 + \sum_{y \mid (x, y) \notin \mathcal{D}} f(x, y)^2, \\
    \Phi_{B}(f) = \mathbb{E}_{\mathcal{D}}\left[ \phi_{B}(f, D) \right] = \sum_{y=1}^N \left( f(x, y) - f_B(x, y) \right)^2.
\end{gather*}

The following theorem guarantees the consistency of this surrogate:

\begin{theorem}
    \label{consistency theorem}
    The surrogate loss $\Phi_{B}$ is $\operatorname{DCG}$-consistent, i.e.,
    \begin{equation}
        \mathcal{L}_{\operatorname{DCG}}(\bm{\pi}) - \mathcal{L}_{\operatorname{DCG}}^* \leq \left(2 \sum_{y=1}^N c_y^2 \right)^{\frac{1}{2}} \left( \Phi_{B}(f) - \Phi_{B}^* \right)^{\frac{1}{2}},
    \end{equation}
    where $\Phi_B^* = 0$ under the achievable assumption.
\end{theorem}

Notably, the ground truth interaction scores $r_{x,y}$ are not restricted to binary values. Since \textbf{DCG} depends only on the rank ordering of the predicted scores, any order-preserving transformation $g(\cdot): \mathbb{R} \to \mathbb{R}$ satisfies $\operatorname{DCG}(f) = \operatorname{DCG}(g(f))$. This leads to the following corollary:

\begin{corollary}
    A squared-form surrogate function of the form
    \begin{equation}
        \phi(f, D) = \sum_{y=1}^N w_x \left( f(x, y) - r_{x,y} \right)^2
    \end{equation}
    is $\operatorname{DCG}$-consistent if:
    \begin{enumerate}
        \item $w_x > 0$,
        \item The scores of positive instances are strictly greater than those of negative instances, i.e., $\min_{(x, y) \in D} r_{x,y} > \max_{(x, y) \notin D} r_{x,y}$.
    \end{enumerate}
\end{corollary}

Based on this corollary, we establish the consistency of the \textbf{RG$^2$} objective with respect to \textbf{DCG} as stated below:

\begin{theorem}
    \label{thm:consistency}
    For any given $\bm{o} \in \mathbb{R}^N$, there exists a constant $C$ such that:
    \begin{equation*}
        \mathcal{L}_{\operatorname{DCG}}(\bm{o}) - \mathcal{L}_{\operatorname{DCG}}^* \leq C \left( \mathcal{L}_{\operatorname{RG^2}}(\bm{o}) - \mathcal{L}_{\operatorname{RG^2}}^* \right)^{\frac{1}{2}}.
    \end{equation*}
\end{theorem}

Furthermore, the consistency of $\mathcal{L}_{\operatorname{RG^2}}$ can also be interpreted through the lens of Bregman Divergence. Specifically, for the loss formulations in Eq.~(\ref{eq.RGsquaredform}):
\begin{align}
    \mathcal{L}_{\operatorname{RG^2}} &= D_{\phi_2}\left( g_2(\bm{o}^{(x)}), \bm{\eta} \right), \\
    \mathcal{L}_{\operatorname{RG^\times}} &= D_{\phi_\times}\left( g_\times(\bm{o}^{(x)}), \bm{\eta} \right),
\end{align}
where the generating functions and link functions are defined as:
\begin{gather*}
        \phi_2(\bm{o}^{(x)}) = |\mathcal{I}_x| \| \bm{o}^{(x)} \|^2, \quad g_2(\bm{o}^{(x)}) = 2|\mathcal{I}_x| \bm{o}^{(x)}, \\
        \phi_\times(\bm{o}^{(x)}) = |\mathcal{I}_x| {\bm{o}^{(x)}}^\top \left( I - \frac{1}{N} \bm{1}_N \bm{1}_N^\top \right) \bm{o}^{(x)}, 
        \quad g_\times(\bm{o}^{(x)}) = 2|\mathcal{I}_x| \left( I - \frac{1}{N} \bm{1}_N \bm{1}_N^\top \right) \bm{o}^{(x)}, \\
        [\bm{\eta}]_y = r_{x,y} \frac{N}{|\mathcal{I}_x|} - 1.
\end{gather*}

It is straightforward to verify that the functions $\phi$ above are convex and differentiable (since $I - \frac{1}{N} \bm{1}_N \bm{1}_N^\top$ is positive semidefinite), and that the corresponding $g$ functions are inverse order-preserving. This establishes that the proposed \textbf{RG} losses are \textbf{DCG-consistent}.

We emphasize that the inverse order-preserving property of $g$ relies on the assumption that the loss function applies uniform weighting across samples, as adopted in Eq.~(\ref{eq.RGsquaredform}). In contrast, assigning different weights to positive and negative samples (as in \textbf{WSL}) may disrupt this property and hence affect \textbf{DCG-consistency}. Nonetheless, in practical scenarios, assigning larger weights to top-ranked positive samples and smaller weights to negative samples often aligns with task objectives. Our analysis provides a theoretical justification for the squared-form loss achieving good performance on ranking metrics despite such weighting heuristics.

\subsection{Instantiation with Matrix Factorization}
In this section, we utilize \textbf{RG} losses in \textbf{MF} model with analysis on a generalization upper bound and optimize it using \textbf{ALS} method. An alternative with the NewtonCG method on DNN-based models is provided in \cite{pu2024learning}.

\textbf{MF} is a typical and flexible collaborative filtering method that has been comprehensively developed in various similarity learning scenarios (\cite{koren2008factorization,hu2008collaborative,koren2009matrix,rendle2012bpr,lian2014geomf}). It decomposes the large interaction matrix $R \in \{0,1\}^{M \times N}$ where $r_{x,y} = 1 $ if $(x,y) \in D$, into two lower-dimensional matrices.

Consider a \textbf{MF}-based ranking task with $M$ left components and $N$ right components. Let $P\in\mathbb{R}^{M\times K}, Q\in\mathbb{R}^{N\times K}$ stand for the representation matrices to be learned, where $K$ stands for the dimension of latent embeddings. The predicted matrix is computed as $O = P\cdot Q^\top$, with an apparent property $\text{rank}(O)\leq K$. The flexibility of \textbf{MF} lies in the selection of different objective functions, which significantly impact the model's performance on representing similarity. By utilizing our proposed \textbf{RG} losses, rewriting Eq.(\ref{eq.RGsquaredform}-\ref{eq.RG+}) into \textbf{MF}-based form with Frobenius-norm regularizers, we have
\begin{equation}\label{eq:RG2 MF form}
\begin{aligned}
    \mathcal{L}_{\text{RG$^2$}} &=\sum_{x,y=1}^{M,N}W_{x,y}(S_{x,y}-P_{x\cdot}\cdot Q_{\cdot y}^\top)^2+\lambda(\|P\|_F^2+\|Q\|_F^2)\\
    \mathcal{L}_{\text{RG$^\times$}} &=\sum_{x,y=1}^{M,N}W_{x,y}(S_{x,y}-P_{x\cdot}\cdot Q_{\cdot y}^\top)^2+\lambda(\|P\|_F^2+\|Q\|_F^2)-\sum_{x=1}^M V_x P_{x\cdot}Q^\top QP_{x\cdot}
\end{aligned}
\end{equation}
where $P_{x\cdot}$ and $Q_{\cdot y}^\top$ denotes the $x$-th row and the $y$-th column of $P$ and $Q^\top$ respectively, and
\begin{equation}\label{eq:def_of_matrices}
S_{x,y} =r_{x,y}\frac{N} {|\mathcal{I}_x|}-1, \quad W_{x,y},V_{x} = \begin{cases}|\mathcal{I}_x|,\frac{|\mathcal{I}_x|}{N}  & \text{following Eq.(\ref{eq.RGsquaredform})}\\ w_{x,y},\frac{|\mathcal{I}_x|(n+1)}{N^2}  & \text{following Eq.(\ref{eq.RG+})}\end{cases}
\end{equation}
Here we fuse the representations of $W_{x,y}$ in Eq. (\ref{eq.RGsquaredform}) and Eq. (\ref{eq.RG+}) for ease of description.

\subsection{Generalization Analysis}
To analyze the generalization performance of the proposed \textbf{RG} losses in matrix factorization (MF), we first define the hypothesis space as follows:
\[
h(x, y) = o_{y}^{(x)} = P_{x\cdot} \cdot Q_{\cdot y}^\top,
\]
where $P \in \mathbb{R}^{M \times K}$ and $Q \in \mathbb{R}^{N \times K}$ are the latent factors with embedding dimension $K$. We assume that the latent factors are constrained by:
\[
\|P_{x\cdot}\|_2 \leq C_P, \quad \|Q_{\cdot y}\|_2 \leq C_Q, \quad \forall x \in X, y \in Y,
\]
which implies:
\[
|h(x, y)| \leq C_P C_Q.
\]
Without loss of generality, we may rescale $h(x, y)$ into the interval $[0, 1]$ via normalization if necessary. The hypothesis space is thus defined as:
\[
\mathcal{H} := \left\{ h_{O} : (x, y) \mapsto P_{x\cdot} Q_{\cdot y}^\top \ \middle|\ \text{rank}(O) \leq K,\ \|P_{x\cdot}\|_2 \leq C_P,\ \|Q_{\cdot y}\|_2 \leq C_Q \right\}.
\]

Correspondingly, the empirical error and generalization error in our settings are formed as:
\begin{equation*}
\begin{aligned}
    \hat{\mathcal{R}}_{\operatorname{RG^2}}(h) &= \frac{1}{|D|} \sum_{x,y=1}^{M,N}|\mathcal{I}_x| \left(o_y^{(x)}+1 - r_{x,y}\frac{N} {|\mathcal{I}_x|} \right)^2\\
     \mathcal{R}_{\operatorname{RG^2}}(h) &= \mathbb{E}_{D\sim\mathcal{D}}\left[|\mathcal{I}_x| \left(o_y^{(x)}+1 - r_{x,y}\frac{N} {|\mathcal{I}_x|} \right)^2\right]\\
    \hat{\mathcal{R}}_{\operatorname{RG^\times}}(h) &= \frac{1}{|D|} \sum_{x,y=1}^{M,N}|\mathcal{I}_x| \left(\left(o_y^{(x)}+1 - r_{x,y}\frac{N} {|\mathcal{I}_x|} \right)^2-\frac{1}{N}(\bm{1}_N^\top\bm{o}^{(x)})^2\right)\\
     \mathcal{R}_{\operatorname{RG^\times}}(h) &= \mathbb{E}_{D\sim\mathcal{D}}\left[|\mathcal{I}_x| \left(o_y^{(x)}+1 - r_{x,y}\frac{N} {|\mathcal{I}_x|} \right)^2-\frac{1}{N}(\bm{1}_N^\top\bm{o}^{(x)})^2\right]
\end{aligned}
\end{equation*}
\begin{remark}
    Both error can be restricted by a $L$-Lipschitz continuity, where $L>2(2\sqrt{N}C_PC_Q+1+N)$.
\end{remark}

The discussion in Appendix.\ref{Appdix.GUpperBound} proposes the following conclusion on generalization upper bound for both \textbf{RG} losses: 
\begin{theorem}\label{Thm:GUpperBound}
    Suppose $|\hat{\mathcal{R}}(h)|\leq B, \forall h\in\mathcal{H}$, then with probability at least $1-\delta$, we have
     \begin{equation}\label{eq.generalization bound}
        \mathcal{R}(h)\leq \hat{\mathcal{R}}(h)+16\sqrt[d+2]{\frac{(d+1)B^4e^{d+1}L^{d}}{|D|\delta}}
     \end{equation}
     where $d = K(M+N)\log\frac{16eM}{K}$.
\end{theorem}

We further combine this upper bound with Theorem~\ref{thm:consistency} to obtain a lower bound for ranking metric \textbf{DCG}.
\begin{proposition}
    \begin{equation}
    \begin{aligned}
        \mathbb{E}_{\mathcal{D}}\left[-\operatorname{DCG}(\bm{o}^{(x)},D)\right]&\leq C(\mathcal{R}(h)-\Phi^*)^{\frac{1}{2}}+\mathcal{L}_{\operatorname{DCG}}^*\\
        &\leq C\left(\hat{\mathcal{R}}(h)+\mathcal{T}_D(d,B,e,\delta)-\Phi^*\right)+\mathcal{L}_{\operatorname{DCG}}^*
    \end{aligned}
    \end{equation}
    Note that $\mathcal{L}_{\operatorname{DCG}}^*=\mathbb{E}_{\mathcal{D}}\left[\operatorname{IDCG}(\bm{o}^{(x)},D)\right]$. Then with elementary transformations, we have
    \begin{equation}
    \begin{aligned}
        \mathbb{E}_{\mathcal{D}}\left[\operatorname{DCG}(\bm{o}^{(x)},D)\right]\geq \mathbb{E}_{\mathcal{D}}\left[\operatorname{IDCG}(\bm{o}^{(x)},D)\right]- C\left(\hat{\mathcal{R}}(h)-\inf_h \mathcal{R}(h)+\mathcal{T}_{D}(d,B,e,\delta)\right)
    \end{aligned}
    \end{equation}
    where $\mathcal{T}_{D}(d,B,e,\delta) = 16\sqrt[d+2]{\frac{(d+1)B^4e^{d+1}L^{d}}{|D|\delta}}$.
\end{proposition}

It is worth noting that the generalization lower bound of \textbf{DCG} derived from this proposition is constrained by the pseudo-dimension $d$ of the hypothesis space and the number of interactions $|D|$ with $O((d+1)^{\frac{1}{d+2}}\cdot C^{1-\frac{1}{d+2}})$ and $O(|D|^{-\frac{1}{d+2}})$, respectively.

\subsubsection{Optimization with ALS}
The proposed \textbf{RG} losses is well-suited for utilization in \textbf{ALS} optimization algorithm. This algorithm updates the model parameters based on the closed-form solutions. Regarding Eq.(\ref{eq:RG2 MF form}), by fixing one of the matrix $P$ or $Q$, the derivative with respect to the other term is calculated as:
\begin{displaymath}
\begin{split}
    \operatorname{RG^2}:
    \frac{\partial\mathcal{L}_{\operatorname{RG^2}}}{\partial P_{x,k}}&=2\sum_y W_{x,y}(P_{x\cdot}Q_{\cdot y}^\top-S_{x,y})Q_{y,k}+2\lambda \left(\sum_y W_{x,y}\right)P_{x,k}\\
    \implies \frac{\partial\mathcal{L}_{\operatorname{RG^2}}}{\partial P_{x\cdot}}&=\left[\frac{\partial\mathcal{L}}{\partial P_{x,1}}, \cdots,\frac{\partial\mathcal{L}}{\partial P_{x, M}}\right]=2P_{x\cdot}\left(Q^\top\widetilde{W}_{x\cdot}Q+\lambda \left(\sum_y W_{x,y}\right)I\right)-2S_{x\cdot}\widetilde{W}_{x\cdot}Q\\
     \frac{\partial\mathcal{L}_{\operatorname{RG^2}}}{\partial Q_{y\cdot}}&=2Q_{\cdot y}\left(P^\top\widetilde{W}_{\cdot y}P+\lambda \left(\sum_{x}W_{x,y}\right)I\right)-2S_{\cdot y}^\top\widetilde{W}_{\cdot y}P\\
    \operatorname{RG^\times}:\frac{\partial \mathcal{L}_{\operatorname{RG^\times}}}{\partial P_{x\cdot}}&=2 P_{x\cdot}\left(Q^{\top} \widetilde{W}_{x\cdot} Q+\lambda\left(\sum_y W_{x,y}\right) I-V_x Q^{\top} Q\right)-2 S_{x\cdot} \widetilde{W}_{x\cdot} Q\\
    \frac{\partial \mathcal{L}_{\operatorname{RG^\times}}}{\partial Q_{y\cdot}}&=2 Q_{y\cdot}\left(P^{\top} \widetilde{W}_{\cdot y} P+\lambda\left(\sum_{x}W_{x,y}\right) I - P^\top \widetilde{V} P \right)-2 S_{\cdot y} \widetilde{W}_{\cdot y} P\\
\end{split}
\end{displaymath}
where $\widetilde{W}_{x\cdot},\widetilde{W}_{\cdot 
 y},\widetilde{V}$ are diagonal matrices with the elements of $W_{x\cdot},W_{\cdot y}, V_x$ on the diagonal. 

Therefore, the objective reaches its minimum at a closed-form solution by setting derivatives to zero. For \textbf{RG$^2$}, 
\begin{align}
\label{UpdateP}P_{x\cdot}&=S_{x\cdot}\widetilde{W}_{x\cdot}Q\left(Q^\top\widetilde{W}_{x\cdot}Q+\lambda\left(\sum_{y}W_{x,y}\right) I\right)^{-1}\\
\label{UpdateQ}
Q_{i\cdot} &= S_{\cdot y}^\top\widetilde{W}_{\cdot y}P\left(P^\top\widetilde{W}_{\cdot y}P+\lambda\left(\sum_{x}W_{x,y}\right) I\right)^{-1}
\end{align}

And for \textbf{RG$^\times$},
\begin{align}
\label{UpdateP*}P_{x\cdot}&=S_{x\cdot}\widetilde{W}_{x\cdot}Q\left(Q^\top\left(\widetilde{W}_{x\cdot}-V_x I\right)Q+\lambda\left(\sum_{y}W_{x,y}\right) I\right)^{-1}\\
\label{UpdateQ*}
Q_{y\cdot} &= S_{\cdot y}^\top\widetilde{W}_{\cdot y}P\left(P^\top\left(\widetilde{W}_{\cdot y}-\widetilde{V}\right)P+\lambda\left(\sum_{x}W_{x,y}\right) I\right)^{-1}
\end{align}
which can be referred to an adjustment to the coefficient of $Q^\top Q$. The equations above serve as the update rule for model parameters as illustrated in Algorithm.\ref{alg:wALS}. First, the matrices $P, Q$ are initialized uniformly in line 3, and the matrices $W, S$ are calculated according to Eq.(\ref{eq:def_of_matrices}) in lines 4-7. After initialization, we alternately update the matrices $P$ and $Q$ using Eq.(\ref{UpdateP}) / Eq.(\ref{UpdateP*}) and Eq.(\ref{UpdateQ}) / Eq.(\ref{UpdateQ*}), respectively, until convergence. 
\begin{algorithm}[tb]
   \caption{Weighted Alternating Least Squares.}
   \label{alg:wALS}
    \begin{algorithmic}[1]
        \STATE {\bfseries Input:} Data Matrix ${R} \in \{0,1\}^{M,N}$
        \STATE {\bfseries Output:} ${P}$ and ${Q}$
        \STATE Initialize ${P}$ and ${Q}$
        \STATE $M, N \gets {R}.shape[0], {R}.shape[1]$
        \STATE $\mathbf{x} \gets \left[\sum R_{x\cdot}\right]_{1\le x\le M}$
        \STATE ${W},{V} \gets $ Eq.(\ref{eq:def_of_matrices})
        \STATE ${S} \gets \frac{1}{N}\text{Diag}(\mathbf{x})^{-1} {R}-1$
       \REPEAT 
            \STATE Update ${P}_{x\cdot}$, $\forall\; x$ with Eq.(\ref{UpdateP}/\ref{UpdateP*})
            \STATE Update ${Q}_{y\cdot}$, $\forall\; y$ with Eq.(\ref{UpdateQ}/\ref{UpdateQ*})
       \UNTIL{convergence}
    \end{algorithmic}
\end{algorithm}

Further, we note that in contrast to \textbf{WSL} or Eq.(\ref{eq.RG+}), which impose different weights for positive and negative samples, our initial setting of \textbf{RG$^2$} imposes weights on all samples that are related to context $x$ only (hence we can replace $W_{x,y}$ with $W_x$ without ambiguity). This is a very good property for \textbf{ALS} which allows us to perform matrix operations rather than updating row-by-row.


\subsubsection{Complexity Analysis}
We give out a brief analysis on the time complexity of Alg.\ref{alg:wALS}. Recall that $P, Q$ are matrices shaped $M\times K$ and $N\times K$, with $K\ll \min\{M, N\}$. Note all users share the same constant matrices $S$ and $W$ and thus they can be pre-computed. 

Regarding \textbf{RG$^2$}, the update procedure of $P_{x\cdot}$ comprises the following matrix operations. The multiplication of $Q^\top \widetilde{W}_{x\cdot}$ has a complexity of ${O}(NK)$, as $\widetilde{W}_{x\cdot}$ is diagonal. The subsequent multiplication with $Q$ yields a complexity of ${O}(NK^2)$, while the matrix inversion contributes $O(K^3)$. $S$ is decomposed into a sparse interaction matrix and an all-one matrix, leading to a complexity of $O(|D|K+MK+NK)$ for computing $S_{x\cdot}\widetilde{W}_{x\cdot}Q$, followed by an additional $O(MK^2)$ for multiplying $S_{x\cdot}\widetilde{W}_{x\cdot}Q$ with the inversion result. The overall complexity for updating the left and right component matrix yields $O(|D|K^2+(M+N)K^3)$. After $T$ iterations, the total complexity achieves $O(T(|D|K^2 + (M+N)K^3))$.

As for \textbf{RG$^\times$}, the primary difference from \textbf{RG$^2$} lies in the inclusion of the second-order interaction terms, which introduce additional computations of $V_x Q^\top Q$ and $P^\top\widetilde{V} P$ in each update step, which requires $O(NK^2)$ and $O(MK^2)$ time, respectively. These terms are shared across all objects and can be pre-computed once at the beginning of each iteration. During the update of $P_{x\cdot}$, the interaction term $V_x Q^\top Q$ is involved, but since $Q^\top Q$ is shared, the per-object update complexity remains $O(NK^2 + K^3)$, consistent with \textbf{RG$^2$}. As for $Q_{y\cdot}$, the update complexity also remains $O(MK^2 + K^3)$. These extra computations only introduce an overhead of $O((M + N)K^2)$ per iteration, which does not change the dominant complexity order. After $T$ iterations, the total complexity achieves $O(T(|D|K^2 + (M + N)K^3))$, which remains practical when the latent dimension $K$ is relatively small.


\subsection{Convergence Rate Analysis of Optimization Algorithms}

To further demonstrate the advantages of applying high-efficiency optimization method including \textbf{ALS} and \textbf{NewtonCG}(\cite{pu2024learning}) than using \textbf{SGD}, we briefly introduce an analysis of the convergence rates among three optimization algorithms employed in our study, inspired by \cite{jain2017non}. 

With regard to the joint variables $(P, Q)$ in \textbf{MF} form, the objective function in Eq.(\ref{eq:RG2 MF form}) is generally non-convex. However, the variables $P$ and $Q$ are marginally convex, ensuring robust convergence guarantees.

\begin{definition}
    (Joint Convexity) A continuously differentiable function in two variables $f :
\mathbb{R}^p \times \mathbb{R}^q \to \mathbb{R}$ is jointly convex if for every $(x_1, y_1), (x_2, y_2) \in \mathbb{R}^p\times \mathbb{R}^q$ we have
\begin{equation}
    f(x_2,y_2)\geq f(x_1,y_1)+\langle\nabla f(x_1,y_1), (x_2,y_2)-(x_1,y_1)\rangle
\end{equation}
where $\nabla f(x_1,y_1)$ is the gradient of $f$ at point $(x_1, y_1)$.
\end{definition}

\begin{definition}
    (Marginal Convexity) A continuously differentiable function in two variables $f :
\mathbb{R}^p \times \mathbb{R}^q \to \mathbb{R}$ is marginally convex in its first variable if for every value of $y\in \mathbb{R}^q$,
the function $f(\cdot,y):\mathbb{R}^p\to \mathbb{R}$ is convex, i.e., 
\begin{equation}
    f(x_2,y)\geq f(x_1,y)+\langle\nabla_x f(x_1,y), (x_2,y)-(x_1,y)\rangle
\end{equation}
where $\nabla_x f(x_1,y)$ is the partial gradient of $f$ at point $(x_1, y)$.
\end{definition}

The following table summarizes the theoretical convergence rates of \textbf{SGD}, \textbf{Newton-CG}, and \textbf{ALS} for non-convex problems, as discussed in Appendix.\ref{Appdix.ConvergRate}:

\begin{table}[h]
    \centering
    \begin{tabular}{|c|c|c|c|}
        \hline
        \textbf{Algorithm} &\textbf{Convergence Rate for Non-convex Problems} \\
        \hline
        SGD & Sublinear Convergence ($\sim\mathcal{O}(1/\sqrt{T})$) \\
        \hline
        Newton-CG & Locally Superlinear Convergence \\
        \hline
        ALS & Linear Convergence ($\sim\mathcal{O}(\rho^T), 0 <\rho < 1$) \\
        \hline
    \end{tabular}
    \caption{Comparison of convergence rates for SGD, Newton-CG, and ALS.}
    \label{tab:convergence_rates}
\end{table}

Overall, the convergence rates of both \textbf{Newton's} method and \textbf{ALS} on the non-convex \textbf{MF} task consistently outperform those of the widely-used \textbf{SGD} methods. This observation further underscores the effectiveness and superiority of our proposed loss functions.


%% file: section/exp.tex
\section{Experimental Results}
We conduct extensive experiments to evaluate the effectiveness and efficiency of our proposed \textbf{RG} loss functions. The experiments are designed to address two key questions: 
\begin{itemize}
    \item whether the \textbf{RG} losses achieve superior similarity learning performance compared to existing baselines;
    \item whether the application of highly-efficient \textbf{ALS} significantly improves computational efficiency over the widely-used \textbf{SGD}.
\end{itemize}  

To this end, we compare the performance of \textbf{RG}s with several state-of-the-art loss functions across multiple datasets, including tasks like recommendation and link prediction. Additionally, we analyze the convergence speed of \textbf{ALS} compared to \textbf{SGD}, providing insights into the trade-offs between accuracy and efficiency.

\subsection{Experimental Settings}
\subsubsection{Dataset and Evaluation}

\textbf{Dataset.} We evaluate our method on four public datasets: Three for recommendations: MovieLens-10M, Amazon-electronics, and Steam Games; One for link prediction: Simple Wiki, collected from different real-world online platforms, involving domains of movies, shopping, games and search engines, which are abbreviated as \textbf{MovieLens}, \textbf{Electronics}, \textbf{Steam} and \textbf{Wiki}, whose detailed descriptions can be found in Appendix.\ref{appdix.data}.

\textbf{Metrics.} We adopt widely used ranking evaluation metrics, Mean Reciprocal Rank with cutoff set as K(\textbf{MRR@K}), Mean Average Precision with cutoff set as K(\textbf{MAP@K}) (for link prediction) and \textbf{NDCG@K} (for recommendations), to measure the ranking performance of different methods, which aligns with the theoretical understandings in previous discussions. The definition of all metrics are given in Def.~\ref{def:DCG&NDCG} with a top-K cutoff. We set K = 10 on all datasets and use \textbf{NDCG@10}/\textbf{MAP@10} as the early stop indicator to demonstrate the broad validity of our loss on ranking metrics for recommendation and link prediction tasks respectively.

\subsubsection{Baselines}
To validate the effectiveness of the proposed novel loss function, we incorporate various types of loss functions for ranking tasks as baseline methods, including sampling-based methods like \textbf{BPR}, \textbf{BCE} and \textbf{SSM}; variant of log Softmax loss like \textbf{Sparsemax}. For recommendation tasks, we also choose sampling-based recommendation objectives like \textbf{UIB} and \textbf{SML} and \textbf{WRMF} optimized with the \textbf{ALS} method. The descriptions of baselines and implementation details can be found in Appendix.\ref{appdix.exp}.
\begin{table*}[htb]\small
    \centering
    \caption{Comparisons of recommendation performance. S-Softmax and Softmax are the abbreviations of Sampled Softmax and Softmax loss functions. \textbf{Bold} and \underline{underline} numbers represent the best and the second-best results respectively.}
    \begin{tabular}{l|c|c|c|c|c|c}
    \toprule
         Method & \multicolumn{2}{c|}{MovieLens} & \multicolumn{2}{c|}{Electronics} & \multicolumn{2}{c}{Steam} \\
         \cmidrule(lr){2-3} \cmidrule(lr){4-5} \cmidrule(lr){6-7} 
         & MRR@10 & NDCG@10 & MRR@10 & NDCG@10 & MRR@10 & NDCG@10 \\ \midrule
         BPR & 0.3476 & 0.2116 & 0.0124 & 0.0154 & 0.0434 & 0.0521 \\
         BCE & 0.3602 & 0.2228 & 0.0098 & 0.0126 & 0.0435 & 0.0532 \\
         SSM & 0.3785 & 0.2378 & 0.0119 & 0.0152 & 0.0466 & 0.0549  \\
         Sparsemax & 0.3177 & 0.1738 & 0.0078 & 0.0100 & 0.0317 & 0.0359  \\
         UIB & 0.3836 & 0.2426 & 0.0100 & 0.0133 & 0.0410 & 0.0506 \\
         SML & 0.3386 & 0.2045 & 0.0076 & 0.0103 & 0.0406 & 0.0482  \\
         WRMF & 0.4475 & 0.2797 & 0.0146 & 0.0178 & 0.0464 & 0.0544 \\
         SM & 0.4487 & 0.2849 & 0.0172 & 0.0212 & 0.0493 & 0.0579 \\
         \textbf{RG$^2$} & \underline{0.4720} & \underline{0.2957} & \underline{0.0186} & \underline{0.0229} & \underline{0.0509} & \underline{0.0586} \\
         \textbf{RG$^\times$} & \textbf{0.4737} & \textbf{0.2975} & \textbf{0.0189} & \textbf{0.0233} & \textbf{0.0513} & \textbf{0.0590} \\
    \bottomrule
    \end{tabular}
    \label{tab:performance}
\end{table*}

\begin{table*}[htb]\small
    \centering
    \caption{Comparison of link prediction performance. T/O represents timeout.}
    \begin{tabular}{l|ccccccc}
    \toprule
    Wiki & BPR & BCE & SSM & Sparsemax & SM & \textbf{RG$^2$} & \textbf{RG$^\times$} \\
    \midrule
    MAP@10  & 0.3691 & 0.4527 & 0.5235 & T/O & \textbf{0.5750} & 0.5525 & \underline{0.5537} \\
    MRR@10  & 0.5291 & 0.5813 & 0.6717 & T/O & \textbf{0.7202} & 0.6988 & \underline{0.6996} \\

    \bottomrule
    \end{tabular}
    \label{tab:performance2}
\end{table*}


\subsection{Overall Recommendation Performance}
To validate the effectiveness of \textbf{RG}s in similarity learning performance, we compare our method with baselines in terms of the ranking metrics in the experiments. The results are presented in Table~\ref{tab:performance} and \ref{tab:performance2}. From the results, we can summarize the following findings.

\textit{Findings 1}. Non-sampling methods show significant enhancement in performance compared to their sampling components. As for recommendation tasks, (non-sampling) \textbf{WRMF}, \textbf{SM} and \textbf{RG}s show significant enhancement compared with (sampling) \textbf{BPR, BCE, SSM, UIB} and \textbf{SML}. And for link prediction, \textbf{SM} and \textbf{RG}s over the other methods. Despite optimized by SGD, \textbf{Softmax} loss demonstrates significant superiority over all sampling baselines, which obtains 17.4\% average relative improvements compared with the best sampling baseline in terms of all metrics and all tasks, which aligns with its theoretical properties in consistency with ranking metrics. 

\textit{Findings 2}. Bias has emerged as a critical factor constraining the performance of all sampling-based and \textbf{SGD}-based methods due to the randomness introduced. For \textbf{ALS}-based non-sampling methods, \textbf{WRMF} and \textbf{RG}s employs a deterministic optimization process, which avoids the variance introduced by \textbf{SGD}, which consistently outperforms other baselines except \textbf{SM}. The best results of \textbf{RG$^\times$} showcases a 6.0\% average relative improvements and a 3.3\% average relative decrease compared with \textbf{SM}, which substantiates the effectiveness of our approximation and the advantage of \textbf{ALS} optimization, which preserves a deterministic updating direction through a closed-form solution.

\textit{Findings 3}. \textbf{RG}s achieves better performance with \textbf{WRMF} (15.3\% at average), demonstrating the better alignment of \textbf{RG}s with ranking metrics within a same \textbf{ALS} optimization method, demonstrating agreements with our theoretical results. \textbf{Sparsemax}, as a sparse variant of \textbf{SM}, sets most probabilities in the output distribution as zero, resulting in the underfitting of representations of inactive users and cold items, thereby suffering severe performance degradation.


\begin{figure*}[!ht]
\vskip -0.2in
    \begin{subfigure}[b]{.5\linewidth}
        \includegraphics[width=\linewidth]{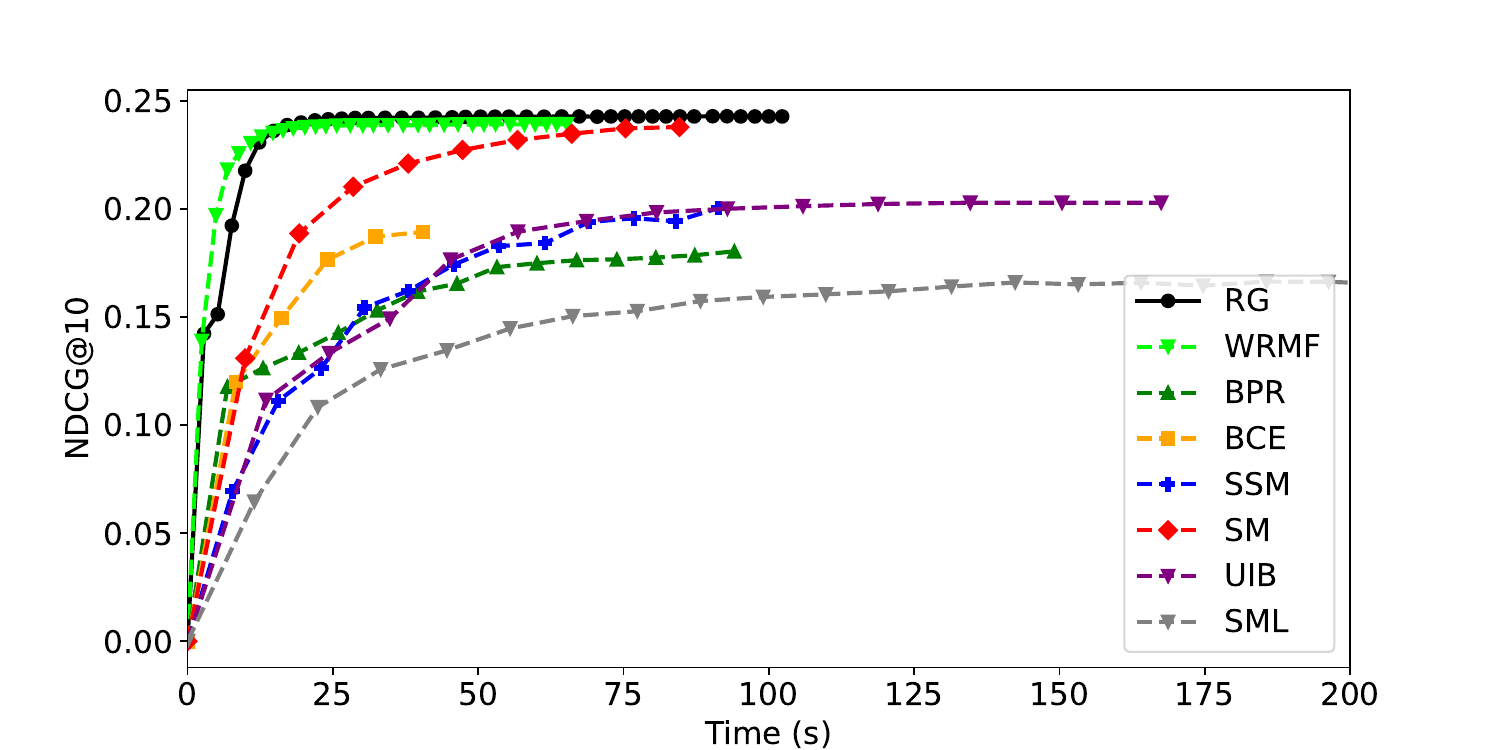}
        \caption{NDCG@10 on MovieLens}
    \end{subfigure}
    \begin{subfigure}[b]{.5\linewidth}
        \includegraphics[width=\linewidth]{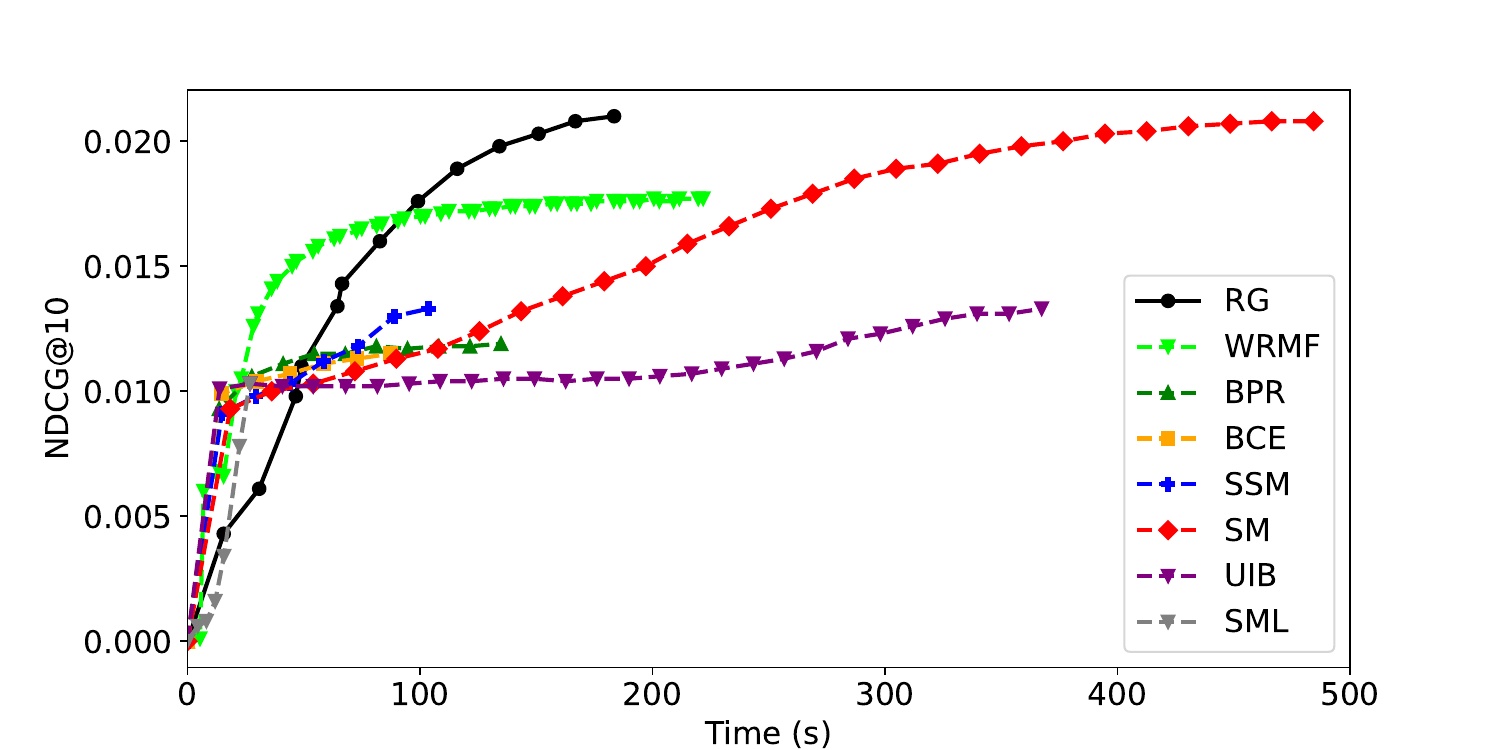}
        \caption{NDCG@10 on Electronics}
    \end{subfigure}
    \begin{subfigure}[b]{.5\linewidth}
        \includegraphics[width=\linewidth]{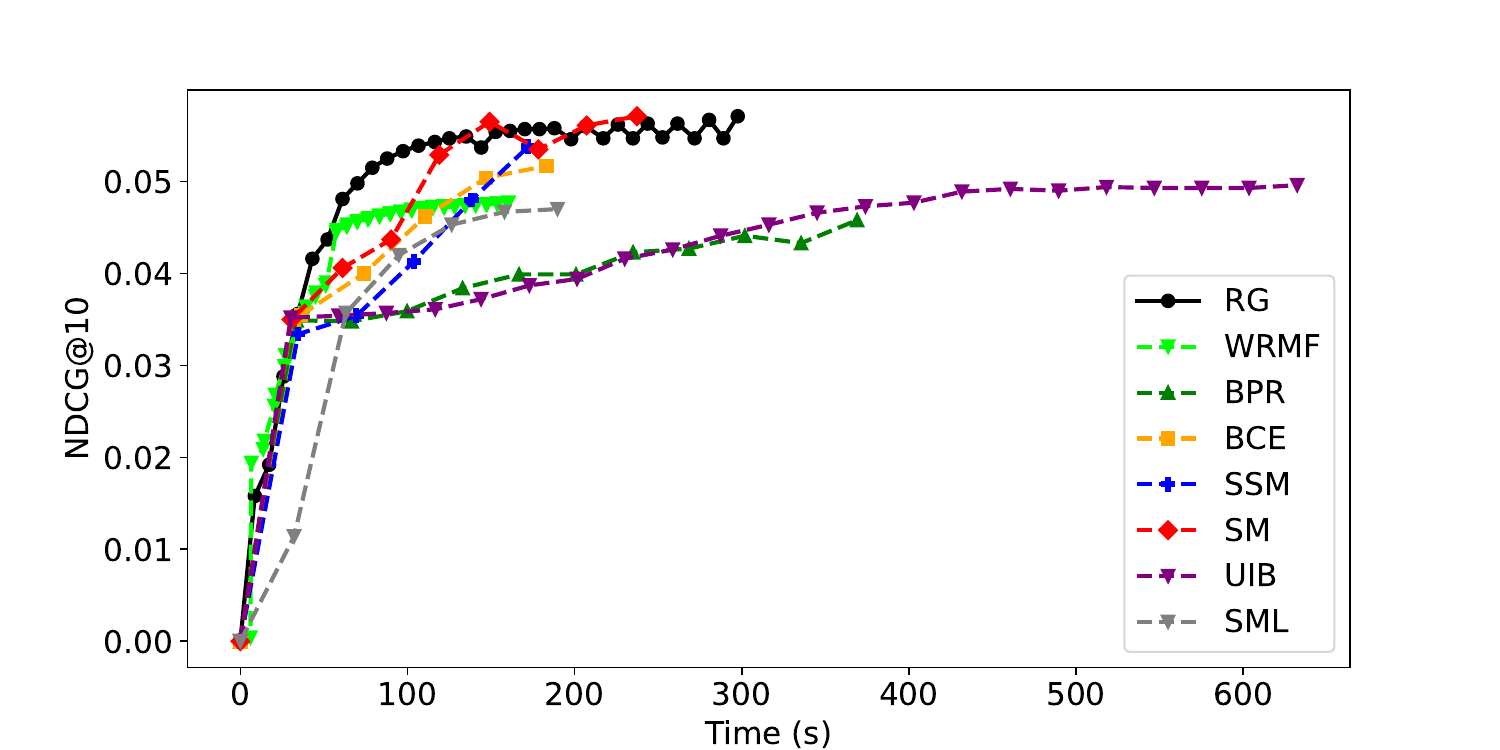}
        \caption{NDCG@10 on Steam}
    \end{subfigure}
    \begin{subfigure}[b]{.5\linewidth}
        \includegraphics[width=\linewidth]{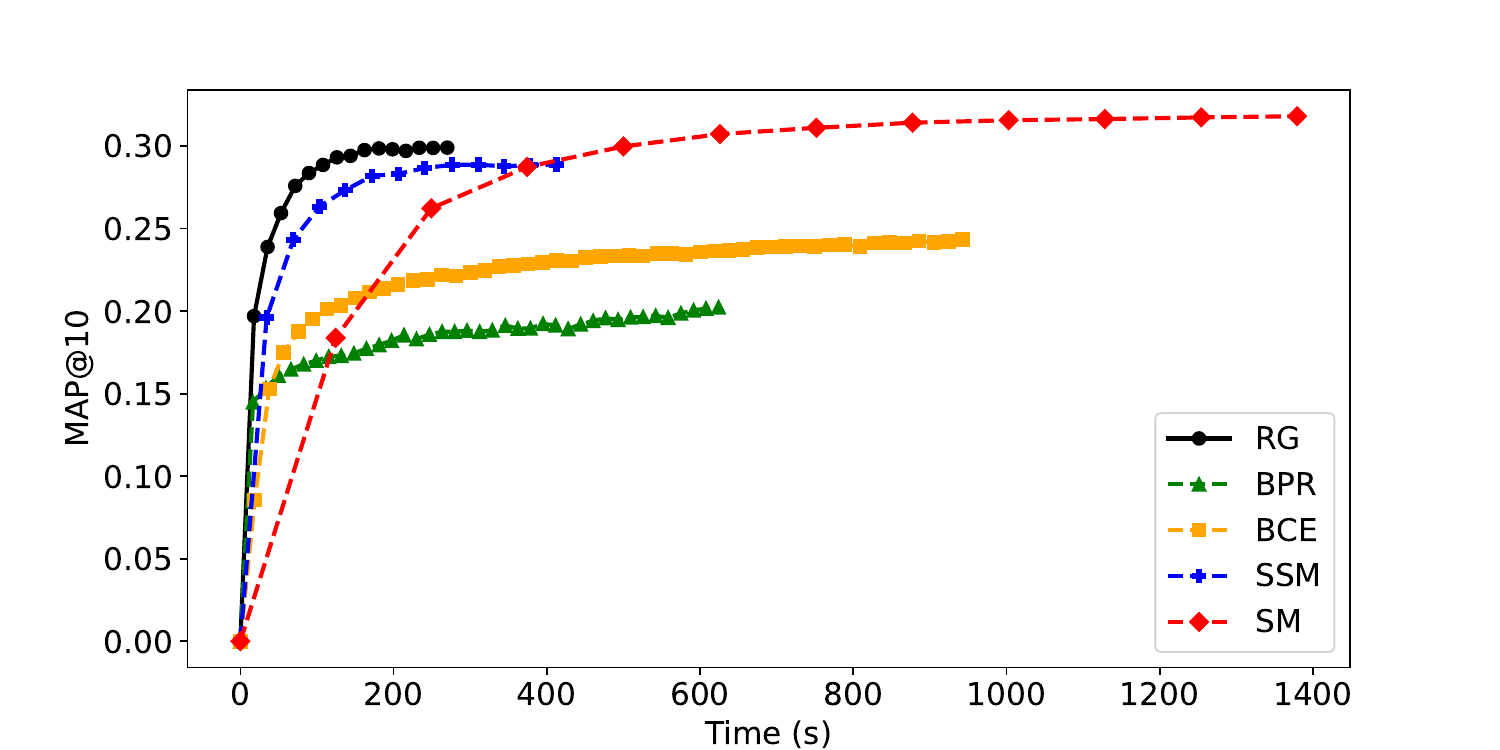}
        \caption{MAP@10 on Wiki}
    \end{subfigure}
\caption{Comparisons of convergence speed on all datasets.}
\label{fig:convergence}
\end{figure*}

\subsection{Comparison of Convergence Speed}\label{sec: convergence}
Furthermore, to investigate the convergence speed of our method, we record the metrics after each epoch evaluated on the validation set during the training process over all recommendation datasets, as illustrated in Figure~\ref{fig:convergence}, where we record the data point of each epoch and their running time. For \textbf{ALS}-based methods (\textbf{RG}\footnote{The data points of \textbf{RG$^2$} and \textbf{RG$^\times$} are very close and hard to distinguish, thus we only retain the results of \textbf{RG$^\times$}.} and \textbf{WRMF}), we merge each 2 epochs (UpdateP and UpdateQ respectively) into 1 point for clarity.

As shown in the figures, both \textbf{RG} and \textbf{WRMF} demonstrate faster convergence speed than \textbf{Softmax}, proving the efficacy of \textbf{ALS} that directly optimizes the convex problem with a closed-form solution. Besides,
sampling-based approaches, such as \textbf{BPR} and \textbf{SampledSoftmax} can reduce their training time of each epoch compared with \textbf{Softmax}, 
especially for datasets with more items, such as \textbf{Electronics} and \textbf{Wiki}. 

%% file: section/conclusion.tex
\section{Conclusion}

In conclusion, our exploration into approximating \textbf{SM} and the introduction of the \textbf{RG} losses mark strides in the domain of extreme similarity learning. The \textbf{RG} losses, ingeniously approximating and upper bounding \textbf{SM/SSM} loss through Taylor expansion, represents an innovative forward in designing learning objectives for ranking similarity learning tasks. This innovation maintains \textbf{RG}'s alignment with \textbf{DCG}, thus ensuring relevance and performance on ranking metrics. Our rigorous empirical analysis, conducted across four public datasets, has confirmed the superiority of the \textbf{RG}s approach. Through the adaptation to ALS optimizations, \textbf{RG} not only achieves performance comparable to the established benchmarks of \textbf{SM} but, in several instances, surpasses them, demonstrating remarkable efficiency improvements in the training process without compromising on performance. Such advancements underscore the potential of newly designed loss functions in enhancing the scalability and effectiveness of extreme similarity learning.